\documentclass[sigconf, nonacm=true]{acmart}
\AtBeginDocument{%
  }
\usepackage[utf8]{inputenc} 
\usepackage[T1]{fontenc}    
\usepackage{hyperref}       
\usepackage{url}            
\usepackage{booktabs}       
\usepackage{amsfonts}       
\usepackage{nicefrac}       
\usepackage{microtype}      
\usepackage{lipsum}
\usepackage{comment}
\usepackage{xcolor}
\usepackage{amsthm}
\usepackage{amsmath}
\usepackage{caption}
\usepackage{subcaption}
\usepackage{multicol} 
\usepackage{graphicx}
\usepackage{natbib}
\usepackage[inline]{enumitem}
\usepackage{multirow}
\usepackage{tcolorbox}
\usepackage{framed} 
\usepackage{subcaption}

\theoremstyle{plain}

\theoremstyle{definition}

\theoremstyle{remark}

\tcbset{
    width=\textwidth,
    colback=white,
    colframe=black,
    fonttitle=\bfseries,
    title=Aspect Summarization Prompt (Cont'd),
    boxrule=0.8mm,
    arc=0mm,
    coltitle=white,
    colbacktitle=black,
    left=2mm,
    right=2mm,
    top=1mm,
    bottom=1mm
}

\setcopyright{acmlicensed}
\copyrightyear{2018}
\acmYear{2018}
\acmDOI{XXXXXXX.XXXXXXX}
\acmConference[Conference acronym 'XX]{Make sure to enter the correct
  conference title from your rights confirmation email}{June 03--05,
  2018}{Woodstock, NY}
\acmISBN{978-1-4503-XXXX-X/2018/06}

\begin{document}

\title{Bridging AI and Science: Implications from a Large-Scale Literature Analysis of AI4Science}

\author{Yutong Xie}
\authornote{These authors contributed equally to this work.}
\email{yutxie@umich.edu}
\affiliation{%
  \institution{University of Michigan}
  \city{Ann Arbor}
  \state{Michigan}
  \country{USA}
}

\author{Yijun Pan}
\authornotemark[1] 
\email{panyijun@umich.edu}
\affiliation{%
  \institution{University of Michigan}
  \city{Ann Arbor}
  \state{Michigan}
  \country{USA}
}

\author{Hua Xu}
\email{hua.xu@yale.edu}
\affiliation{%
  \institution{Yale University}
  \city{New Haven}
  \state{Connecticut}
  \country{USA}
}

\author{Qiaozhu Mei}
\email{qmei@umich.com}
\affiliation{%
  \institution{University of Michigan}
  \city{Ann Arbor}
  \state{Michigan}
  \country{USA}
}

\renewcommand{\shortauthors}{Xie et al.}
\begin{abstract}
Artificial Intelligence has proven to be a transformative tool for advancing scientific research across a wide range of disciplines. However, a significant gap still exists between AI and scientific communities, limiting the full potential of AI methods in driving broad scientific discovery. Existing efforts in identifying and bridging this gap have often relied on qualitative examination of small samples of literature, offering a limited perspective on the broader AI4Science landscape. 
In this work, we present a large-scale analysis of the AI4Science literature, starting by using large language models to identify scientific problems and AI methods in publications from top science and AI venues. Leveraging this new dataset, we quantitatively highlight key disparities between AI methods and scientific problems, revealing substantial opportunities for deeper AI integration across scientific disciplines. Furthermore, we explore the potential and challenges of facilitating collaboration between AI and scientific communities through the lens of link prediction.
Our findings and tools aim to promote more impactful interdisciplinary collaborations and accelerate scientific discovery through deeper and broader AI integration. Our code and dataset are available at: \url{https://github.com/charles-pyj/Bridging-AI-and-Science}. 
\end{abstract}
\vspace{-10pt}


\vspace{-10pt}
\maketitle

\section{Introduction}

The 2024 Nobel Prizes in Physics and Chemistry were both awarded to Artificial Intelligence (AI) researchers. In particular, the developers of AlphaFold \cite{jumper2021highly} were recognized for their groundbreaking work, which is based on the Transformer network and has revolutionized the prediction of protein structures, providing a transformative approach to solving complex biological problems. This compellingly exemplifies how AI has emerged as a powerful tool to facilitate scientific research across various disciplines \cite{wang2023scientific,leontidis2024science,mak2023artificial,guo2021artificial,xu2024ai}. 

However, despite the transformative potential of using AI methods for solving scientific problems (AI4Science), a considerable gap persists between AI and scientific communities, hindering the full exploitation of AI for scientific discovery: On the one side, advanced AI methodologies might remain mysterious or underutilized by scientists; On the other side, AI researchers may lack awareness of the specific challenges and potential applications in scientific domains, missing opportunities for interdisciplinary collaborations.

These challenges invoke a few key research questions: What does the AI4Science landscape look like? Are AI methods and scientific problems evenly interconnected, or are some of them underexplored? And how could we expand the connections?

Existing efforts to approach these questions have predominantly involved small-scale, qualitative reviews of AI's application in science \cite{wang2023scientific,leontidis2024science},  particularly in specialized areas such as drug discovery and material science \cite{mak2023artificial,guo2021artificial,xu2024ai}.
These qualitative analyses of the literature often rely on heuristic insights from domain experts to suggest potential uses of AI in solving scientific problems. While valuable, such focused reviews are limited in providing a comprehensive and diverse perspective on the AI4Science landscape.

In the mean time, recent research from the Science of Science community has initialized larger-scale studies that quantify AI's impact on publications in scientific fields \cite{gao2023quantifying,duede2024oil}, although they often heavily rely on established scientific taxonomies in ``AI-heavy'' fields. There remains a lack of delivering a holistic, dynamic, and data-driven overview of AI4Science. Such analysis is crucial for understanding the barriers 
and identifying new opportunities for deep engagement of fast-evolving AI methodology in scientific research.

Aiming at addressing these key questions, we conduct a large-scale and comprehensive analysis of relevant literature over the past decade. 
We start by using large language models (LLMs) to identify the scientific problems and AI methods addressed in publications from top science and AI venues, through which we assemble a novel and balanced AI4Science dataset to analyze the role of AI in scientific research (Sec. \ref{sec:dataset}). 
Leveraging this dataset, we illustrate the AI4Science landscape through two projection maps (Fig. \ref{fig:landscape}) and a bipartite graph (Fig. \ref{fig:bipartite}a).  We further quantitatively reveal the current discrepancies between AI and science, leading to several key findings and implications (Sec. \ref{sec:gap}). 
Finally, through the lens of link prediction, we demonstrate the potential and challenges in advancing the bridge between AI and science (Sec. \ref{sec:model}). 
Our findings and tools may foster more effective interdisciplinary collaborations and accelerate scientific discovery through a more diverse and deeper integration of AI methods.

\begin{figure*}[htbp]
    \centering
    \includegraphics[width=\linewidth]{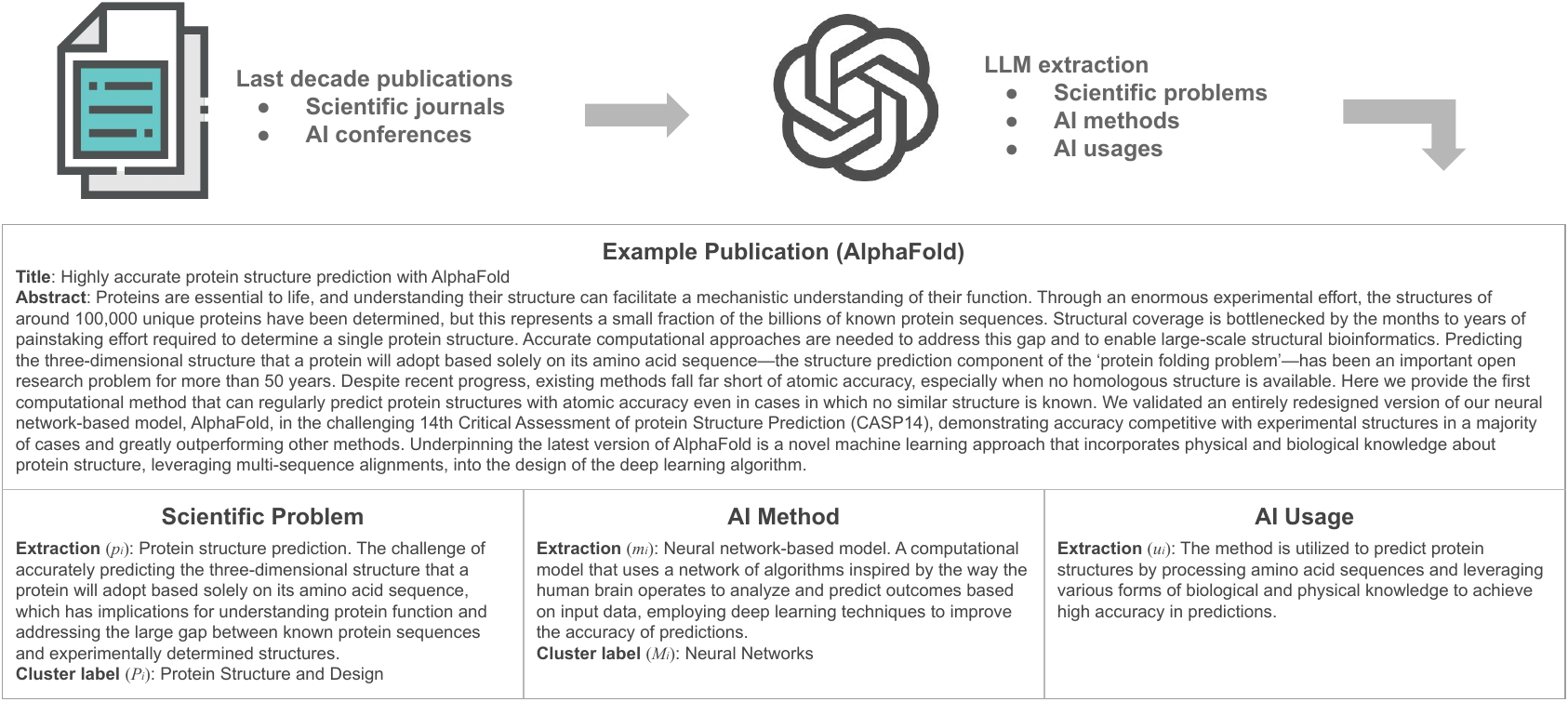}
    \vspace{-20pt}
    \caption{
    Illustration of LLM-based extraction of scientific problem $p_i$, AI method $m_i$, and AI usage $u_i$ from an example publication. The extractions are then semantically projected, clustered, and labeled to form the AI4Science landscape (Fig. \ref{fig:landscape}), as well as to construct the bipartite graph (Fig. \ref{fig:bipartite}a). 
    }
    \vspace{-5pt}
    \label{fig:extraction}
\end{figure*}

\section{A New Dataset and the AI4Science Landscape}
\label{sec:dataset}

A major challenge in understanding the disparities and potential of AI4Science is the lack of a comprehensive and balanced dataset. Existing literature data are often biased toward successful stories, such as areas where AI and science are already deeply intertwined. Or they reflect the perspective of either the scientific or the AI communities, but not both.  
We take the initiative to create a broad and balanced large-scale literature dataset that provides a wide-angled lens for understanding AI4Science research.

\subsection{Curation of AI4Science Literature}

We collect publications from leading science and AI venues to offer a balanced view from both communities: 

\vspace{-3pt}
\begin{itemize}[leftmargin=*]
    \item For science domains, we include three top multidisciplinary journals that represent cutting-edge and high-quality research (\emph{Nature}, \emph{Science}, and \emph{PNAS}) and two of their subjournals (\emph{Nature Communications} and \emph{Science Advances}).
    \item For AI communities, we include seven top conferences from the list of \href{https://csrankings.org/}{CSRankings.org}, comprising two AI-focused venues (\emph{AAAI}, \emph{IJCAI}), three machine learning venues (\emph{ICLR}, \emph{ICML}, \emph{NeurIPS}), and two venues of applied AI research (\emph{SIGKDD}, \emph{WWW}).
\end{itemize}
\vspace{-2pt}

Our data includes a total of 162,656 publications from 2014 to 2024. Appendix \ref{app:data-curation} provides details of the data curation process.

\begin{figure*}[htbp]
    \centering
    \vspace{-5pt}
    \includegraphics[width=\linewidth]{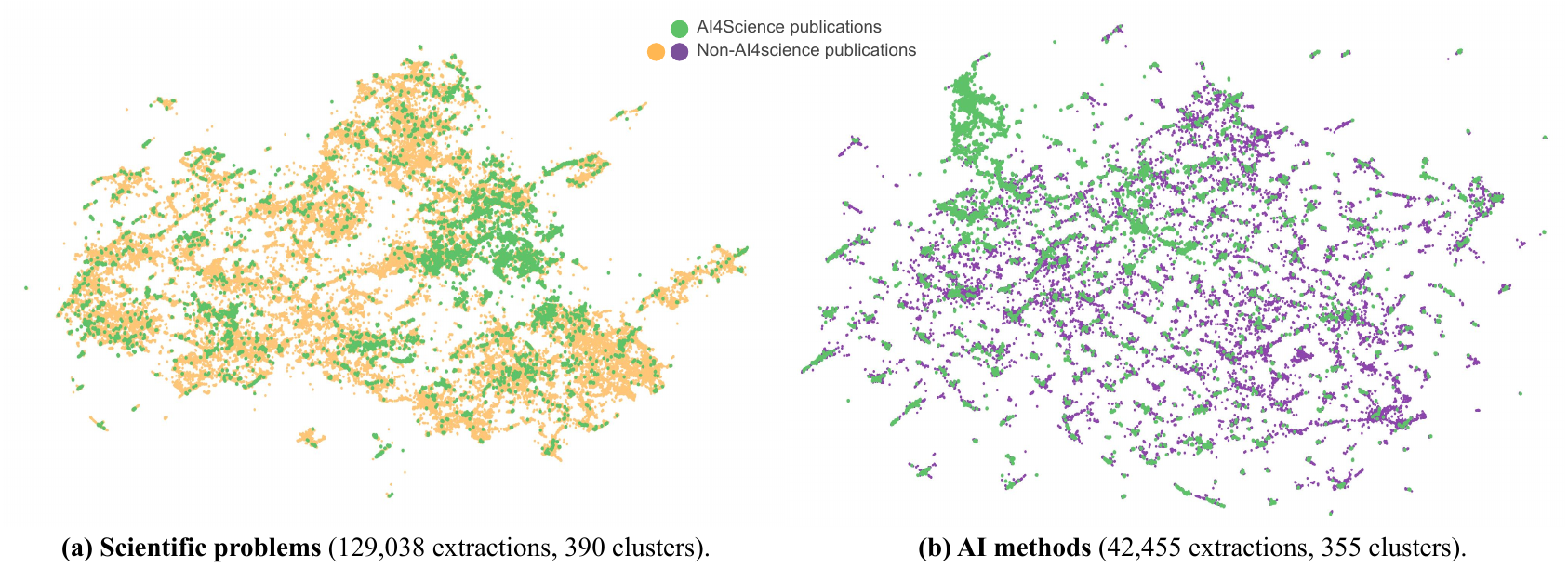}
    \vspace{-20pt}
    \caption{
    2D projection maps of the AI4Science landscape: (a) the extracted scientific problems $\{p_i\}$ and (b) AI methods $\{m_i\}$. Each dot represents a publication, with coordinates obtained by reducing the dimensionality of the problem/method embeddings. 
    \textcolor[HTML]{2ecc71}{\textbf{Green}} dots correspond to \emph{AI4Science} work, using AI methods to address scientific problems; 
    \textcolor[HTML]{FFC300}{\textbf{Orange}} dots show papers addressing scientific problems without using AI methods;
    \textcolor[HTML]{8E44AD}{\textbf{Purple}} dots are papers using AI to address non-scientific problems. 
    The visualizations reveal a noticeable discrepancy in the distribution of \emph{AI4Science} work (\textcolor[HTML]{2ecc71}{\textbf{green}}) versus \emph{non-AI4Science} work (\textcolor[HTML]{FFC300}{\textbf{orange}} and \textcolor[HTML]{8E44AD}{\textbf{purple}}) in both the problem and method spaces.
    }
    \vspace{-5pt}
    \label{fig:landscape}
\end{figure*}

\subsection{Extracting Scientific Problems and AI Methods with LLMs}

To understand how AI is applied to science, we need to extract paired entries of scientific problems and AI solutions from relevant publications. Traditionally, this is accomplished through literature reviews and surveys which involve manual examinations of small samples of papers \cite{wang2023scientific,leontidis2024science}, a process that is labor-intensive and time-consuming. Recent quantitative studies in the Science of Science field have started using metadata entries, such as keywords, to identify AI usage and categorize scientific topics \cite{gao2023quantifying,duede2024oil}. While this method enables larger-scale analysis, it offers only a coarse-grained view, potentially missing the nuanced details of scientific problems and AI methods. Additionally, it can suffer from issues such as incomplete data entries, limiting the effectiveness of analysis.

In contrast, our work leverages large language models (LLMs) to extract detailed descriptions of scientific problems and AI methods from publication titles and abstracts, allowing for a more nuanced, scalable, and data-driven analysis. Such LLM-based extraction techniques have also been employed in recent studies for extracting aspects from scientific literature, such as \citet{zhang2024massw}.

Particularly, for each publication of interest, we use OpenAI GPT-4o mini to extract the following three key aspects:
\vspace{-3pt}
\begin{itemize} 
    \item \textbf{Scientific Problem} ($p_i$): The primary scientific problem addressed in the paper, including both a keyword or keyphrase summarizing the problem and a detailed definition. 
    \item \textbf{AI Method} ($m_i$): The main AI method applied in the paper, including a keyword or keyphrase summarizing the method and a detailed description. 
    \item \textbf{AI Usage} ($u_i$): A detailed explanation of how the AI method is specifically applied to the scientific problem. 
\end{itemize}
\vspace{-3pt}

It is possible that a publication did not address a scientific problem or did not use an AI method, in which case the corresponding field(s) would remain empty. If a paper addresses a scientific problem using an AI method -- i.e., $p_i$, $m_i$, and $u_i$ are all non-empty -- it is considered as an \emph{AI4Science} work. Fig. \ref{fig:extraction} presents an example of the extraction results of an AI4Science publication.

To assess the reliability of GPT extractions, we conducted a small-scale human evaluation on 100 papers. Each extraction record was reviewed by at least two annotators, resulting in an average accuracy of 91.0\%. For more details on the extraction prompts, additional examples of the extracted data, and verification with human annotations, please refer to Appendix \ref{app:data-extract}. 

\subsection{Semantic Projection and Clustering}

One of the key advantages of this new AI4Science dataset compared to the data used in existing studies \cite{gao2023quantifying,duede2024oil} is that our dataset contains detailed textual descriptions of the identified scientific problems and AI methods rather than pre-defined taxonomies, allowing for a deeper semantic analysis of their relationships. 

To achieve this, we generate semantic embeddings for the scientific problems $\{p_i\}$ and AI methods $\{m_i\}$ using the InstructorEmbedding model \cite{su2022one}. We then apply LargeVis \cite{tang2016visualizing} to project these high-dimensional embeddings into 2D coordinates.
Next, a density-based clustering method, HDBSCAN \cite{campello2013density} is utilized to group similar scientific problems and AI methods in this 2D space. For each cluster, we use GPT-4o to generate an understandable summarization based on frequent keywords and samples of extraction results.
This process results in 390 and 355 clusters respectively, with corresponding cluster labels $P_i$ and $M_i$ to each data point.
The choice of the methods and the details of the process are in Appendix \ref{app:data-cluster}. 

Fig. \ref{fig:landscape} shows the 2D maps that illustrate the AI4Science landscape. Fig. \ref{fig:science_all}-\ref{fig:AI_all} in Appendix present the maps with the cluster labels. 

Unlike using keywords or categorizations from publication metadata \cite{gao2023quantifying,duede2024oil}, our approach employs a data-driven methodology. By using textual embedding, clustering, and topic labeling, relevant AI methods discussed in both scientific and AI literature are grouped together, as are the scientific problems mentioned in both fields. This enables a semantic understanding of the AI4Science space and ensures greater consistency and generalizability of the data.

\subsection{Dataset Overview and the Bipartite Graph}

Through LLM-based extraction and semantic clustering, we form a comprehensive dataset, $\mathcal{D}=\{(p_i, P_i, m_i, M_i, u_i)\}_{i=1}^N$, which includes the scientific problems, AI methods, their corresponding cluster labels, and the usage of AI for science. The basic statistics of this dataset are listed in Table \ref{tab:statistics}.

To better illustrate the connections between AI and science, we construct a bipartite graph based on the clusters. In this bipartite graph, scientific problem clusters and AI method clusters are represented as two types of nodes, and the publications act as edges connecting them. Formally, the graph is defined as $\mathcal{G} = (\mathcal{V}, \mathcal{E})$, where $\mathcal{V} = \{P_i\} \cup \{M_i\}$ and $\mathcal{E} = \mathcal{D}$. As shown in Fig. \ref{fig:bipartite}a, the graph provides an intuitive visual representation of how AI is linked to scientific challenges and vice versa.

\begin{table}[t]
    \centering
    \begin{tabular}{l c r}
        \toprule
        \textbf{} & \textbf{Notation} & \textbf{Value} \\
        \midrule
        Publication year range & -- & 2014--2024 \\ 
        \# of venues & -- & 12 \\
        \# of publications & $N$ & 162,656 \\ 
        \# of scientific problem extractions & $\vert\{p_i\}\vert$ & 129,038\\ 
        \# of scientific problem clusters & $\vert\{P_i\}\vert$ & 390 \\ 
        \# of AI method extractions & $\vert\{m_i\}\vert$ & 42,455 \\ 
        \# of AI method clusters & $\vert\{M_i\}\vert$ & 355 \\  
        \# of AI usage extractions & $\vert\{u_i\}\vert$ & 42,455 \\ 
        \# of AI4Science publications & -- & 7,542 \\
        \bottomrule
    \end{tabular}
    \caption{AI4Science dataset $\mathcal{D}=\{(p_i,P_i,m_i,M_i,u_i)\}_{i=1}^N$. 
    }
    \vspace{-10pt}
    \label{tab:statistics} 
\end{table}

\begin{figure*}[htbp]
    \centering
    \begin{subfigure}[htbp]{0.48\textwidth}
        \centering
        \includegraphics[width=\linewidth]{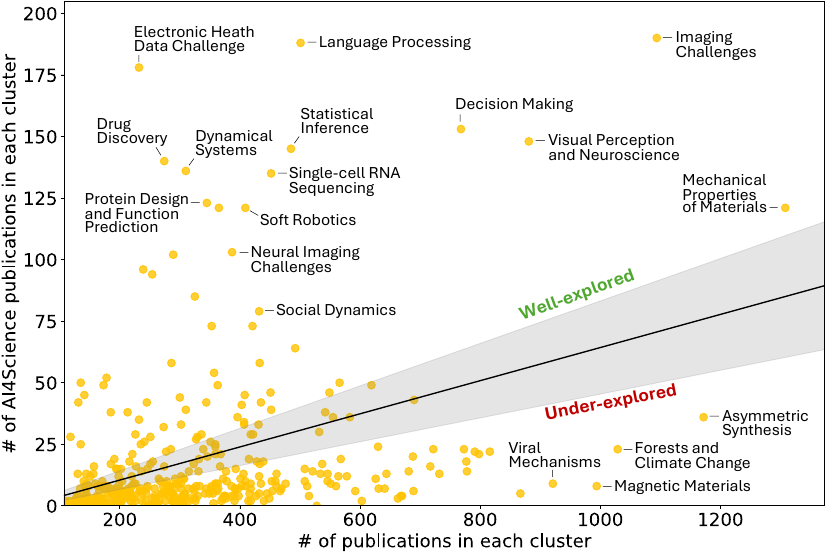}
        \caption{\textcolor[HTML]{FFC300}{\textbf{Scientific problem}} clusters.}
    \end{subfigure}
    \hfill
    \begin{subfigure}[htbp]{0.48\textwidth}
        \centering
        \includegraphics[width=\linewidth]{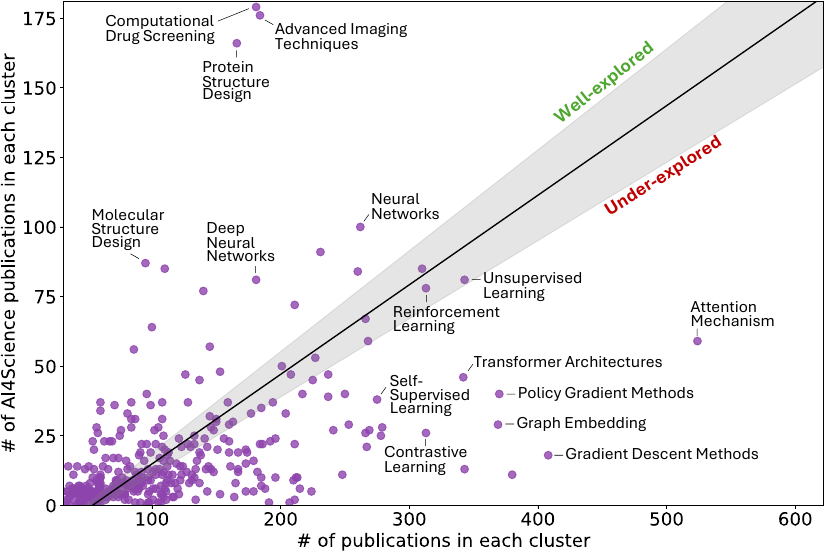}
        \caption{\textcolor[HTML]{8E44AD}{\textbf{AI method}} clusters. }
    \end{subfigure}
    \vspace{-10pt}
    \caption{
    \textcolor[HTML]{FFC300}{\textbf{Scientific problem}} clusters (a) and \textcolor[HTML]{8E44AD}{\textbf{AI method}} clusters (b) are visualized as scatters. 
    In both plots, the $x$-axis represents the total number of publications in each cluster, while the $y$-axis reflects the number of interdisciplinary AI4Science publications in each cluster.
    The black lines show the regression results on the clusters. Clusters above the line indicate regions \emph{well-explored} for AI4Science. Clusters falling below the line highlight the \emph{under-explored} regions, where the integration of AI and science remains limited. These areas represent potential opportunities for further interdisciplinary collaboration. 
    }
    \label{fig:cluster-sizes}
\end{figure*}

\section{Discrepancy Between AI and Science}
\label{sec:gap}

\begin{figure*}[htbp]
    \centering
    \vspace{-5pt}
    \includegraphics[width=\linewidth]{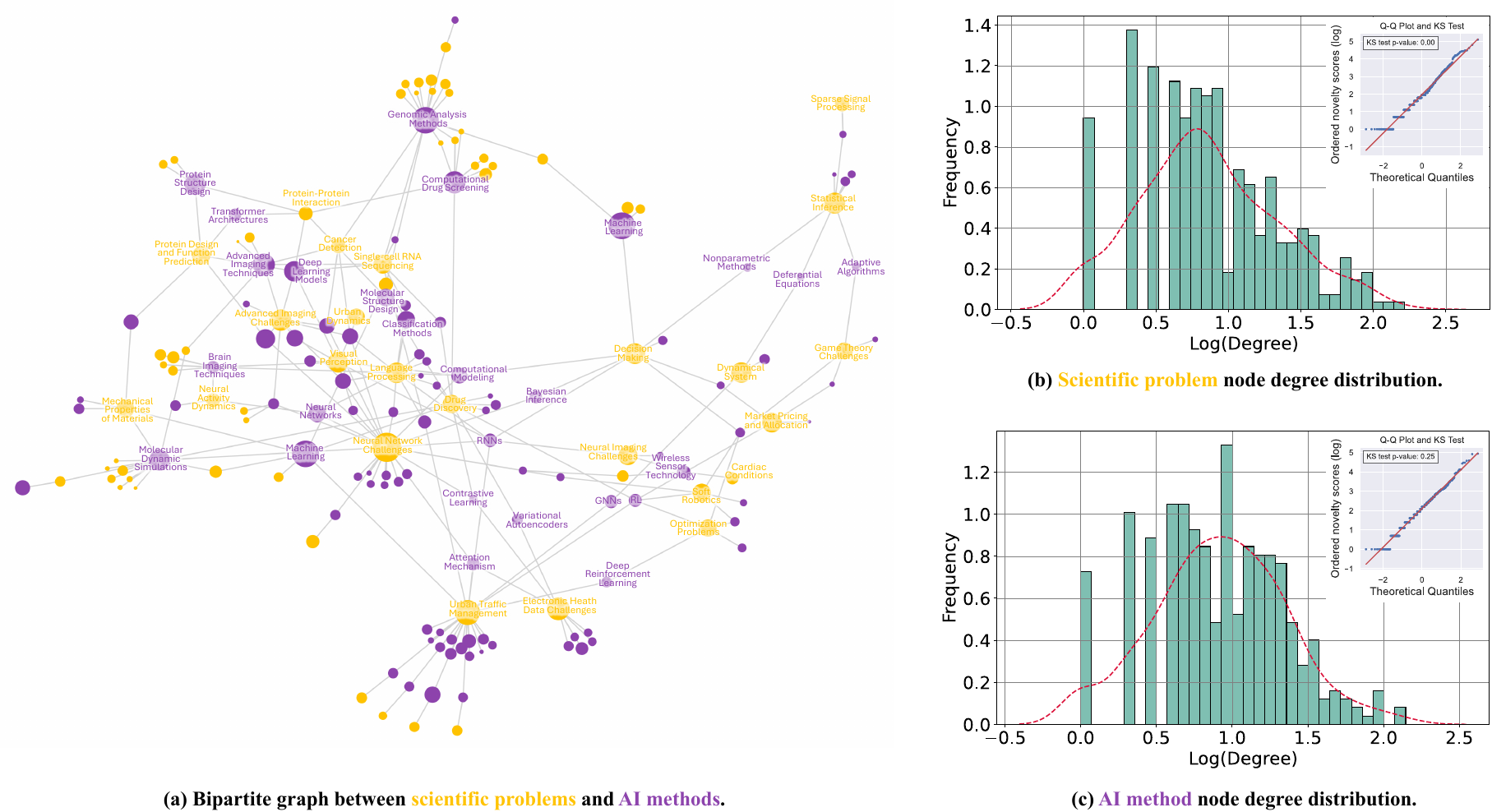}
    \vspace{-18pt}
    \caption{
    The AI-Science bipartite graph and node distributions. 
    (a) The bipartite graph with \textcolor[HTML]{FFC300}{\textbf{scientific problem}} clusters and \textcolor[HTML]{8E44AD}{\textbf{AI method}} clusters as nodes. 
    The size of each node corresponds to its unweighted degree, representing the number of AI methods applied to a scientific problem, or the number of scientific problems utilizing an AI method. For visualization clarity, edges representing fewer than four publications and the resulting isolated noes are hidden. 
    The distribution of links 
    indicates imbalanced connectivity. 
    (b-c) The degree of AI method nodes follows a log-normal distribution, and the degree distribution of scientific problem nodes is even more heavily tailed, suggesting the existence of ``hubs'' in linking AI and science.
    }
    \label{fig:bipartite}
\end{figure*}

Both Fig. \ref{fig:landscape} and \ref{fig:bipartite}a show a noticeable pattern of discrepancy between AI and science. This section quantitatively assesses these disparities, providing a deeper understanding of the gaps. 

\subsection{Uneven Distribution of AI4Science Research}
\label{sec:dis-cluster}

The projection maps in Fig. \ref{fig:landscape} illustrate the distribution of AI4Science work within the semantic spaces of scientific problems and AI methods (the \textcolor[HTML]{2ecc71}{\textbf{green}} dots among \textcolor[HTML]{FFC300}{\textbf{orange}} and \textcolor[HTML]{8E44AD}{\textbf{purple}} dots). The uneven distributions suggest that AI4Science work is heavily clustered in certain subareas, while it remains underrepresented in wide ranges of both the problem and the method spaces. 

In Fig. \ref{fig:cluster-sizes}, we plot the clusters of scientific problems and AI methods, showing the relation between cluster sizes and the presence of AI4Science publications within each cluster. The slopes of the regression lines represent the average proportions of AI4Science work in the literature of each scientific problem cluster or AI method cluster. Clusters below the regression line indicate problems or AI methods in which the integration of AI4Science is underrepresented (noted the \emph{under-explored} regions), suggesting potential opportunities for further interdisciplinary collaboration.

\footnotetext[1]{Six outstanding data points are skipped from the scatter plots for visualization clarity:
`Neural Network Challenges' (820,375) and `Urban Traffic Management' (551,298) in (a),
`Machine Learning' (770,539), `Deep Learning Models' (605,344),
`Genomic Analysis Methods' (292,287), and `Molecular Dynamics Simulations' (214,210) in (b).}

\vspace{-5pt}
\paragraph{Underexplored scientific problems. }

In the bottom-right section of Fig. \ref{fig:cluster-sizes}a, below the regression line, several scientific problem clusters are highlighted as being rarely approached through AI techniques. Notable examples include
`Asymmetric Synthesis', `Magnetic Materials and Spintronics', `Forests and Climate Change', `Viral Mechanisms', `Evolutionary Paleontology', `Nanostructure Synthesis', `Diabetes and Insulin Regulation', `Bacterial Infections and Immune Response', `Cancer Signaling Pathways', and `Cancer Drug Resistance'. 
Scientific problems in this region could benefit from more engagement of AI methods. 

\vspace{-5pt}
\paragraph{Underutilized AI methods. }

Similarly, Fig. \ref{fig:cluster-sizes}b reveals AI techniques that have not yet been widely applied to sciences, including: 
`Attention Mechanisms', `Gradient-Based Methods', `Graph Embedding', `Regularization Methods', `Transformer Architectures', and `Contrastive Learning'. 
These techniques could have been utilized in a wider range of scientific applications.

\subsection{Uneven Distribution of Bipartite Links}
\label{sec:degree}

Fig. \ref{fig:bipartite} reveals an uneven distribution of node sizes in the bipartite graph, indicating the presence of ``hub'' and ``peripheral'' nodes among both scientific problems and AI methods. This implies that certain scientific challenges and AI techniques play a central role in interdisciplinary AI4Science work, while others are less explored.

Quantitatively, \emph{heavy-tailed} degree distributions are observed as visualized in Fig. \ref{fig:bipartite}(b-c). 
Specifically, the degrees for AI method nodes follow a log-normal-like distribution, while the degree distribution for scientific problem nodes exhibits an even heavier tail, though not as extreme as a power-law distribution. 
These findings support the idea that a small number of ``hubs'' are linked to a large variety of AI methods or scientific problems, while many other nodes remain peripheral to these interdisciplinary connections. Heavy-tailed distributions, particularly log-normals, are also commonly observed in other bipartite networks \cite{vasques2018degree}, such as document-term graphs \cite{wang2014discover}, and in collaborative networks \cite{perc2010growth, sun2014analyzing}. This suggests that the structure of the AI4Science bipartite graph reflects a general pattern of interdisciplinary and collaborative work.

Table \ref{tab:degree-highest} in Appendix lists the ``hubs'' with the highest node degrees, representing the most interconnected scientific problems and AI methods.
From the table, we observe that the scientific problems linked to the widest variety of AI methods are primarily computational or data-processing problems, such as: 
`Neural Network Challenges' and `Statistical Inference'. 
These are followed by more domain-specific problems like: 
`Urban Traffic Management', `Electronic Health Data Challenges', `Visual Perception and Neuroscience', `Market Pricing and Allocation', `Single-cell RNA Sequencing', `Soft Robotics', and `Drug Discovery'. 

On the AI methods side, the techniques applied to the greatest number of scientific challenges include general methods such as:
`Machine Learning', `Deep learning Models', `Neural Networks', `Classification Methods', `Data Analysis'.
In addition, there are specialized techniques tailored to scientific domains, including:
`Genomic Analysis Methods', `Computational Drug Screening', `Molecular Dynamics Simulations', `Protein Structure Design', and `Molecular Structural Design'.

\begin{table}[t]
    \centering
    \vspace{3pt}
    \begin{tabular}{l|l|r}
        \toprule
        \textbf{Node type} & \textbf{Partition} & \textbf{Avg. Deg.} \\
        \midrule 
        \multirow{2}{*}{\raggedright \shortstack{Scientific problem\\clusters (n=390)}} 
        & All nodes &  13.0\\
        & Well-explored  & 41.3  \\
        & Under-explored  & 5.0 \\
        \hline
        \multirow{2}{*}{\raggedright \shortstack{AI method\\clusters (n=355)}} 
        & All nodes & 14.3\\
        & Well-explored & 19.1\\
        & Under-explored & 8.1 \\
        \bottomrule
    \end{tabular}
    \caption{Average degree of nodes in the bipartite graph. 
    }
    \vspace{-10pt}
    \label{tab:avg-deg}
\end{table}

We further break down the node degree distribution based on cluster partitions. Specifically, we calculate the average node degree for well-explored and under-explored clusters, as identified earlier in Sec. \ref{sec:dis-cluster}. The statistics are listed in Table \ref{tab:avg-deg}. We observe a higher average degree for well-explored scientific problems and AI methods. 
This suggests that under-explored regions not only have fewer AI4Science publications, but also exhibit a lesser variety, or limited connectivity in the bipartite graph. 

\subsection{Discrepancy Between the AI and Science Communities}
\label{sec:communities}

Our dataset includes publications from both scientific journals and AI conferences, allowing us to explore how these two communities approach the integration of AI into science in distinct ways.

Table \ref{tab:communities} in Appendix lists the top clusters emphasized by each community. 
In terms of facilitating scientific discovery with AI, the science community places greater emphasis on challenges such as: 
`Single-cell RNA Sequencing', `Mechanical Properties of Materials', `Protein Design and Function Prediction', `Cancer Detection', and `Genetic Variants and Traits';
In contrast, the AI community tends to focus more on areas like: 
`Urban Traffic Management', `Electronic Health Data Challenges', `Statistical Inference', `Market Pricing and Allocation', `Soft Robotics', etc. 

When it comes to applying AI methods to scientific problems, the two communities also show different preferences. In addition to general approaches like 
`Machine Learning' and `Deep Learning Models', the science community frequently employs AI techniques specifically tailored for scientific challenges, such as:
`Genomic Analysis Methods', `Molecular Dynamics Simulations', and `Computational Drug Screening'. 
In comparison, the AI community utilizes a broader range of general AI methodologies, covering diverse techniques such as: 
`Reinforcement Learning', `Attention Mechanisms', `Policy Gradient Methods', `Multitask Learning', `Spatio-Temporal Analysis', `Causal Inference', `Self-Supervised Learning', and `Unsupervised Learning'. 

In addition, by comparing Table \ref{tab:communities} with Table \ref{tab:degree-highest}, we also find that the scientific problems addressed by the AI community, tend to be connected to a wider variety of AI methods. Similarly, AI methods frequently used by the science community are applied across a broader range of scientific challenges.
Similarly, AI methods frequently used by the science community are also more likely to be applied across a broader range of scientific challenges. 
This highlights the discrepancy between the two communities, where the science community engages with a limited range of AI methods compared to the AI community, and vice versa.

The distribution of AI4Science work published in scientific journals and AI conferences in the semantic maps (Fig. \ref{fig:combined_distribution} in Appendix) also illustrates the different focuses of these two communities.

\subsection{Findings and Implications}

Through our analysis of the distribution of AI4Science work, bipartite graph node degrees, and the discrepancies between the AI and science communities, we derive the following key findings:

\begin{enumerate}[label=(F\arabic*),leftmargin=*]
    \item Different AI and science subdomains exhibit varying degrees of engagement in AI4Science research, leaving a substantial number of scientific problems and AI methods underexplored in the collaboration context (Sec. \ref{sec:dis-cluster});
    \item The connectivity between scientific problems and AI methods is highly imbalanced, with certain nodes acting as ``hubs'' while other peripheral nodes are less connected (Sec. \ref{sec:degree});
    \item The science and AI communities take distinct approaches to integrate AI into scientific research, prioritizing different problems and methods (Sec. \ref{sec:communities}). 
\end{enumerate}

These key findings lead to two important implications for fostering a wider and deeper exploration of AI4Science:

\begin{enumerate}[label=(I\arabic*),leftmargin=*]
    \item More attention should be directed toward exploring the under-explored areas, incorporating a broader range of scientific challenges and AI techniques into the AI4Science landscape. 
    \item Efforts should be made to discover new connections between AI and science, uncovering innovative ways to apply AI methodologies to scientific research. 
\end{enumerate} 

In the following section, we present preliminary explorations of addressing these implications through link prediction. 

\section{Bridging AI and Science: \\Through the Lens of Link Prediction}
\label{sec:model}

Building on the findings and implications above, we dive deeper into the potentials and challenges of advancing the connections between AI and science from the perspective of link prediction.

\begin{table*}[htbp]
\centering
\begin{tabular}{|l|l|cccc|cccc|cccc|}
\hline
\textbf{Setting} & \textbf{Method} & \multicolumn{4}{c|}{\textbf{Precision}} & \multicolumn{4}{c|}{\textbf{Recall}} & \multicolumn{4}{c|}{\textbf{F1}} \\
& & @1 & @3 & @5 & @10 & @1 & @3 & @5 & @10 & @1 & @3 & @5 & @10 \\ \hline
\multirow{3}{*}{Sci $\to$ AI} & Katz index & 0.115 & 0.108 & 0.086 & 0.076 & 0.015 & 0.035 & 0.052 & 0.104 & 0.026 & 0.053 & 0.065 & 0.088 \\ 
 & Node2vec & 0.198 & 0.185 & 0.168 & 0.129 & 0.033 & 0.098 & \textbf{0.138} & \textbf{0.197} & 0.056 & 0.128 & 0.151 & 0.156  \\
& LLM (Cluster) & \textbf{0.352} & \textbf{0.300} & \textbf{0.238} & \textbf{0.159}  & 0.053 & 
0.101 & 0.134 & 0.176 & 0.093 & \textbf{0.151} & \textbf{0.171} & \textbf{0.167}  \\ 
 & LLM (Paper)  & 0.282 & 0.227 & 0.200 & 0.156  & \textbf{0.058} & \textbf{0.106} & 0.129 & 0.155 & \textbf{0.096} & 0.145 & 0.157 & 0.155 \\ 

\hline
\multirow{3}{*}{AI $\to$ Sci} & Katz index & 0.094 & 0.120 & 0.110 & 0.102 & 0.010 & 0.028 & 0.041 & 0.079 & 0.018 & 0.045 & 0.060 & 0.089 \\
 & Node2vec & 0.096 & 0.108 & 0.110 & 0.109 & 0.013 & 
0.042 & 0.058 & 0.104 & 0.023 & 0.060 & 0.076 & 0.106 \\
& LLM (Cluster)  & 0.201 & 0.168 & 0.143 & 0.129 & 0.016 & 0.042 & 0.053 & 0.092 & 0.030 & 0.068 & 0.077 & 0.107 \\ 
 & LLM (Paper)  & \textbf{0.377} & \textbf{0.323} & \textbf{0.307} & \textbf{0.296} & \textbf{0.065} & \textbf{0.104} & \textbf{0.127} & \textbf{0.163} & \textbf{0.111} & \textbf{0.158} & \textbf{0.180} & \textbf{0.211} \\ 
\hline
\end{tabular}
\caption{Link prediction results of different models. \textbf{Bold} numbers highlights the highest performances in each scenario. 
}
\vspace{-18pt}
\label{tab:results-all}
\end{table*}

\begin{table*}[htbp]
\centering
\begin{small}
\begin{tabular}{|c|c|c|cccc|cccc|cccc|}
\hline
\multicolumn{2}{|c|}{\textbf{Link prediction}} & &
\multicolumn{4}{c|}{\textbf{Precision}} &
\multicolumn{4}{c|}{\textbf{Recall}} &
\multicolumn{4}{c|}{\textbf{F1}} \\
\cline{1-2}
\multicolumn{1}{|c|}{\textbf{Setting}} & \textbf{Method} &
\textbf{Region} & @1 & @3 & @5 & @10 & @1 & @3 & @5 & @10 & @1 & @3 & @5 & @10 \\ 
\hline

\multirow{4}{*}{\shortstack{Sci $\to$ AI}} 
  & \multirow{2}{*}{LLM (Paper)}
    & Well-explored   
      & 0.293 & 0.259 & 0.230 & 0.184 
      & 0.025 & 0.059 & 0.073 & 0.105
      & 0.046 & 0.095 & 0.111 & 0.134 \\
  & 
    & Under-explored 
      & 0.250 & 0.195 & 0.171 & 0.129
      & 0.111 & 0.192 & 0.224 & 0.247
      & 0.153 & 0.193 & 0.194 & 0.169 \\
\cline{2-15}
  & \multirow{2}{*}{LLM (Cluster)}
    & Well-explored
      & 0.360 & 0.411 & 0.353 & 0.228
      & 0.030 & 0.082 & 0.123 & 0.164
      & 0.055 & 0.137 & 0.182 & 0.191 \\
  &
    & Under-explored
      & 0.194 & 0.125 & 0.089 & 0.051
      & 0.068 & 0.122 & 0.147 & 0.182
      & 0.101 & 0.124 & 0.111 & 0.079 \\
\hline

\multirow{4}{*}{\shortstack{AI $\to$ Sci}}
  & \multirow{2}{*}{LLM (Paper)}
    & Well-explored
      & 0.403 & 0.340 & 0.334 & 0.332
      & 0.045 & 0.078 & 0.096 & 0.126
      & 0.080 & 0.127 & 0.149 & 0.182 \\
  &
    & Under-explored
      & 0.364 & 0.301 & 0.267 & 0.236
      & 0.151 & 0.222 & 0.264 & 0.311
      & 0.214 & 0.255 & 0.265 & 0.269 \\
\cline{2-15}
  & \multirow{2}{*}{LLM (Cluster)}
    & Well-explored
      & 0.328 & 0.251 & 0.183 & 0.185
      & 0.020 & 0.047 & 0.053 & 0.102
      & 0.038 & 0.079 & 0.082 & 0.132 \\
  &
    & Under-explored
      & 0.076 & 0.057 & 0.074 & 0.102
      & 0.025 & 0.051 & 0.051 & 0.037
      & 0.038 & 0.054 & 0.060 & 0.055 \\
\hline

\end{tabular}
\end{small}
\caption{Link prediction results on \emph{well-} and \emph{under-explored} regions. }
\label{tab:results-explore}
\end{table*}

\subsection{Methodology}

In the curated AI4Science dataset $\mathcal{D} = \{(p_i, P_i, m_i, M_i, u_i)\}_{i=1}^N$, the extracted and clustered scientific problems $\{p_i\}$ and $\{P_i\}$, as well as AI methods $\{m_i\}$ and $\{M_i\}$, represent the respective landscapes of these two domains. The AI4Science publications, along with their descriptions of AI usage $\{u_i\}$, highlight the connections between scientific challenges and AI methods. To model these connections, we employ the link prediction formulation.

\vspace{-5pt}
\paragraph{Data. }
We split the data into training and test sets based on publication dates. The training set includes papers published between 2014 and 2022 (6287 AI4Science publications), which are used for training models, retrieving data points for generation augmentation, and identifying well/under-explored areas, as illustrated in Fig. \ref{fig:cluster-sizes}.
The testing set consists of publications from 2023 and 2024 (1225 AI4Science publications), serving as the ground truth for evaluating link prediction models. 
The statistics are provided in Table \ref{tab:train-test-data} in Appendix.
Note that the analyses prior to this section involve the full AI4Science dataset, spanning the period from 2014 to 2024.

\vspace{-5pt}
\paragraph{Model and evaluation. }
We employ two types of models to predict links between scientific problems and AI methods. The first category directly make predictions on the \emph{cluster level} with conventional bipartite link prediction models \cite{ozer2024link}, such as those based on the Katz index \cite{katz1953new} and node2vec embeddings \cite{grover2016node2vec} or large language models (LLMs). Formally, given the cluster-level bipartite graph constructed on the training data and a source node $P_i$ (or $M_i$), the model predicts $K$ target nodes that could be potentially linked to the source node. We then compare the predicted nodes with the ground-truth target node set in the test data and report Precision, Recall, and F1 scores @K \cite{al2006link, yang2015evaluating}.
Note that existing links may also appear in the ground-truth set, because a new AI4Science work may repeat an existing connection between a scientific problem and an AI method. Specifically, when using LLM for link prediction, the bipartite links to the source nodes are provided in prompts, and the model is referred to as `LLM (Cluster)'. 

In addition, we also utilize LLMs for \emph{paper-level} generative predictions. Specifically, given a scientific problem $p_i$ from cluster $P_i$, we prompt the LLM to generate potential AI methods $m_i$ that could be applied. These generated method descriptions are then mapped to cluster labels $M_i$ through embedding similarity. Similarly, given an AI method $m_i$, we prompt the LLM to generate potential scientific problems $p_i$ that it could address. The prediction process is further enhanced by retrieving similar papers from the training set using retrieval-augmented generation (RAG) \cite{lewis2020retrieval}.
The model is referred to as `LLM (Paper)'. 

We use OpenAI’s \texttt{gpt-4o-2024-08-06} as the LLM model and report its performance in the main text. Additionally, we also tested the \texttt{gpt-3.5-turbo-0125} model as the generator, given that its knowledge is limited to September 2021, ensuring no risk of data leakage. Our results show that the performance of LLM (Paper) remains consistent, suggesting a low risk of data leakage. However, the performance of LLM (Cluster) declines, likely due to GPT-3.5’s limited ability of reasoning and processing long contexts.
For more detailed descriptions of the models and evaluation metrics, please refer to Appendix \ref{app:experiment}.

\subsection{Overall Link Prediction Results}

Table \ref{tab:results-all} presents the models’ performances on the test data, where LLM-based link prediction methods outperform conventional approaches in most settings, highlighting their strong potential for predicting AI4Science research directions.

Notably, LLM (Cluster) performs better in the Sci$\rightarrow$AI setting (predicting AI solutions for scientific problems), while LLM (Paper) excels in the AI$\rightarrow$Sci setting. This contrast suggests two complementary strategies for conducting AI4Science research:
(1) Given a scientific problem, broadly exploring AI techniques at a higher level can help identify potential solutions. 
(2) Given an AI method, starting from the method itself and its relevant publications can help determine which scientific problems it  could be applied to. 

For conventional link prediction methods, while both Katz and node2vec rely on suggesting similar nodes to link to, node2vec outperforms Katz, suggesting the underlying mechanism connecting scientific problems and AI methods goes beyond homophily in the local neighborhood and potentially considers the broader structures of the AI4Science network. This finding further highlights the potential utility of the AI4Science bipartite graph.

\subsection{Well- and Under-Explored Regions}

Fig. \ref{fig:cluster-sizes} reveals the presence of well- and under-explored areas in both the scientific problem and AI method spaces. Echoing Implication (I1), we comparatively explore the potentials and challenges in facilitating interdisciplinary work for these regions. 

Formally, we refer to scientific problem clusters or AI method clusters above the regression lines as \emph{well-explored} areas, while those that fall below the lines are the \emph{under-explored} regions. Clusters that reside within the 95\% confidence intervals of the regression results are excluded from this analysis. Note that for this experiment, we use only the publications in the training split to decide the well- and under-explored regions. 

Table \ref{tab:results-explore} compares the link prediction results for well-explored and under-explored clusters. In both categories of link prediction methods, we observe a similar trend: Precision scores are generally higher for well-explored scientific problems and AI methods, even as K increases. 
In contrast, Recall is relatively higher for under-explored clusters. 
These results are consistent with the node degree statistics in Table \ref{tab:avg-deg}, where the average degree of under-explored nodes is much lower, making it easier for the models to predict links, resulting in a higher Recall. 

Additionally, in the under-explored partition, the rapid decline in Precision as K increases suggests that link prediction models are capable of discovering novel links that have not been actually explored by researchers in the test period. This highlights great opportunities for using ``AI'' to facilitate the expansion of the connections between AI methods and scientific challenges.

\begin{table}[t]
    \centering
    \vspace{5pt}
    \begin{tabular}{l|rrrr}
        \toprule
        \textbf{Method} & \multicolumn{4}{|c}{\textbf{\# of novel (unseen) links}} \\
        & @1 & @3 & @5 & @10 \\
        \midrule
        Node2vec & 8 & 132 & 432 & 1623 \\
        LLM (Cluster) & 243 & 867 & 1,523 & 3,209 \\
        LLM (Cluster)$^*$ & 430 & 1,650 & 2,999 & 6,963 \\
        LLM (Paper) & 639 & 2,291 & 3,771 & 8,276 \\
        \bottomrule
    \end{tabular}
    \caption{Number of novel links discovered by models compared with the training set (data up to 2022). 
    The publications during 2023 and 2024 introduced 683 new links. For a fair comparison, in LLM (Cluster)$^*$ we control the number of API calls to match that of LLM (Paper).
    }
    \vspace{-23pt}
    \label{tab:new-links}
\end{table}

\subsection{Discovering Novel Links}

Inspired by the capability of link prediction models to uncover novel links, we further explore the newly discovered connections 
in response to Implication (I2).

Compared to the training data up to 2022, publications during 2023 and 2024 actually identified 683 new links between scientific problems and AI methods. 
In contrast, link prediction methods exhibit the potential to discover even more connections, especially as K increases, as shown in Table \ref{tab:new-links}.
Remarkably, the LLM-based approaches exhibit strong potential in generating new links, even with a relatively small number of predictions per data point (e.g., @1). This observation aligns with recent research suggesting that LLMs are capable of proposing novel research ideas \cite{si2024can}. 
Interestingly, LLM (Paper) demonstrates a stronger capability in discovering novel links compared to LLM (Cluster), even when the number of API calls is controlled. This may be due to the greater diversity of paper-level inputs, leading to more varied outputs. 

Many of the predicted links have already materialized into publications, such as applying `Graph Neural Networks' to `Protein Design and Function Prediction' \cite{meller2023predicting}, using algorithms for `Molecular Dynamics Simulations' to tackle challenges in `Polymer Dynamics and Chirality' \cite{kuenneth2023polybert} and `Earth's Mantle Dynamics' \cite{doi:10.1126/sciadv.adh3784}, or leveraging `Multimodal Learning' for `Visual Perception and Neuroscience' \cite{choi2023dualstreamneuralnetworkexplains}. 
Link prediction also highlights numerous unexplored but promising directions, such as using `Reinforcement Learning' for `Carbon Emissions Mitigation', applying `Graph Neural Networks' to predict `RNA-Protein Interactions', and using `Explainable AI' to address challenges in `Artificial Intelligence Ethics`.

\section{Related Work} 

\subsection{Qualitative Literature Reviews of AI4Science}

To identify potential applications of AI techniques in science, researchers typically turn to qualitative literature reviews of relevant papers. These reviews include general explorations of AI's usage across various scientific domains \cite{wang2023scientific,leontidis2024science}, as well as more focused examinations in specific fields such as drug discovery \cite{dara2022machine,mak2023artificial,blanco2023role,qureshi2023ai}, materials design \cite{sha2020artificial,li2020ai,guo2021artificial}, and social sciences \cite{xu2024ai}.

While these reviews provide valuable insights, they often rely on interdisciplinary experts to manually identify and categorize scientific problems and potential AI methods, limiting their scope. Such an approach often could not offer a comprehensive and data-driven overview of the broader AI4Science landscape.

\vspace{-5pt}
\subsection{Quantitative Literature Analysis}

In addition to heuristic and qualitative literature reviews, a few recent papers from the Science of Science community have focused on quantitatively analyzing the AI4Science publications. For instance, \citet{gao2023quantifying} discovered that AI use is prevalent across various scientific disciplines, and papers incorporating AI tend to receive a citation advantage. \citet{duede2024oil} explored AI's engagement in different fields, uncovering an ``oil-and-water'' phenomenon -- while AI-engaged work is spreading across disciplines, it does not integrate well with other work within each specific field.

Although these studies provide valuable insights into AI4Science, they primarily focus on areas where AI is already applied to the scientific problems, overlooking the broader and underexplored parts of the landscape. Moreover, these approaches often rely on pre-defined keywords or n-grams to tag AI4Science research, leading to a taxonomy that provides only a coarse representation of the AI and science space. This method fails to capture the actual semantics and nuances of the language, ultimately limiting the depth of analysis.

By contrast, our study provides a more comprehensive view by analyzing literature from both AI and scientific venues over the past decade, encompassing not only AI4Science but also non-AI4Science publications. We leverage LLMs to extract detailed textual descriptions, enabling a nuanced and scalable data-driven categorization that facilitates deeper analysis. Our depiction of the landscape, quantitative analysis of discrepancies, and link prediction experiments offer fresh insights into the challenges and opportunities in bridging AI and science.

\section{Limitations, Challenges, and Future Work}

\paragraph{Potential bias in data selection. }
As an emerging and interdisciplinary field, cutting-edge and influential AI4Science studies often originate from prominent venues of general interest. For instance, groundbreaking work such as the AlphaFold~\cite{jumper2021highly} was published in Nature. Based on this observation, we have focused on top-tier venues from the last decade. 
While this ensures high-quality and impactful contributions, it could introduce a selection bias that may overlook significant contributions from less prominent venues, potentially resulting in an incomplete representation of the AI4Science landscape. 
Given the flexibility of our analysis framework, researchers can integrate additional data sources from less-prominent venues to achieve a more representative view.

\vspace{-5pt}
\paragraph{LLM-based extraction from titles and abstracts.}
The LLM-based extraction in this study relies solely on paper titles and abstracts, due to limited access to full texts, challenge of processing PDF formats, relatively low signal-to-noise ratio in full texts, and cost considerations. 
Our approach is comparable to the existing AI4Science literature analysis efforts, which also focus on the title and abstract as in the metadata \cite{gao2023quantifying,duede2024oil}. 
While key research problems and methods are often described in abstracts, and human evaluation confirms the accuracy of LLM extraction (Appendix \ref{app:data-extract}), we may miss details presented in the full text of publications. 
Future work can build upon our general and flexible framework to incorporate full-text analysis, enriching the extracted content and capturing more nuanced AI methodologies used in scientific research.

\vspace{-5pt}
\paragraph{Evaluation of link prediction. }
Another limitation of our study is the reliance on publication data as the ground truth for evaluating link prediction models. While this provides a reliable benchmark, it may not fully capture the breadth of potential AI4Science connections, and overlook novel or unconventional links that the models might suggest. 
Future work could consider incorporating more comprehensive evaluation strategies, such as expert reviews or real-world validation, to assess the effectiveness of the link predictions.

\vspace{-5pt}
\paragraph{Limited exploration in link prediction models. }
Our goal is not to optimize the link prediction model, but rather to use link prediction as a lens to demonstrate the potential of connecting AI methods and scientific problems. As a result, we only included a few straightforward link prediction models, leaving room for further research to improve the performance of AI4Science link prediction or building recommender systems for AI methods or scientific problems. 

\section{Conclusion}

This paper aims to depict the AI4Science landscape, identify current gaps between AI and science, while exploring potential ways to bridge them. We introduce a comprehensive, large-scale dataset of curated AI4Science publications, with scientific problems and AI methods extracted using large language models. Through quantitative analysis of this dataset, we uncover several key disparities:
\begin{enumerate*}
    \item Different AI and science subdomains exhibit varying degrees of engagement in AI4Science research, leaving a substantial number of scientific problems and AI methods underexplored;
    \item The connectivity between scientific problems and AI methods is highly imbalanced, with certain nodes acting as ``hubs'' while other peripheral nodes are less connected;
    \item The science and AI communities take distinct approaches to integrating AI into scientific research, prioritizing different problems and methods. 
\end{enumerate*}
We further investigate the potential and challenges of fostering AI4Science collaboration through the lens of link prediction. The experiment results demonstrate great opportunities for using link prediction models to explore the under-investigated scientific problems and AI methods, as well as discover novel connections. 
We anticipate that our findings and tools will provide valuable insights to enhance interdisciplinary collaboration and accelerate scientific discovery through the integration of AI.

\newpage
~\indent
\newpage
\bibliographystyle{ACM-Reference-Format}
\bibliography{sample-base}


\begin{thebibliography}{37}


\ifx \showCODEN    \undefined \def \showCODEN     #1{\unskip}     \fi
\ifx \showISBNx    \undefined \def \showISBNx     #1{\unskip}     \fi
\ifx \showISBNxiii \undefined \def \showISBNxiii  #1{\unskip}     \fi
\ifx \showISSN     \undefined \def \showISSN      #1{\unskip}     \fi
\ifx \showLCCN     \undefined \def \showLCCN      #1{\unskip}     \fi
\ifx \shownote     \undefined \def \shownote      #1{#1}          \fi
\ifx \showarticletitle \undefined \def \showarticletitle #1{#1}   \fi
\ifx \showURL      \undefined \def \showURL       {\relax}        \fi
\providecommand\bibfield[2]{#2}
\providecommand\bibinfo[2]{#2}
\providecommand\natexlab[1]{#1}
\providecommand\showeprint[2][]{arXiv:#2}

\bibitem[Al~Hasan et~al\mbox{.}(2006)]%
        {al2006link}
\bibfield{author}{\bibinfo{person}{Mohammad Al~Hasan}, \bibinfo{person}{Vineet Chaoji}, \bibinfo{person}{Saeed Salem}, {and} \bibinfo{person}{Mohammed Zaki}.} \bibinfo{year}{2006}\natexlab{}.
\newblock \showarticletitle{Link prediction using supervised learning}. In \bibinfo{booktitle}{\emph{SDM06: workshop on link analysis, counter-terrorism and security}}, Vol.~\bibinfo{volume}{30}. \bibinfo{pages}{798--805}.
\newblock


\bibitem[Blanco-Gonzalez et~al\mbox{.}(2023)]%
        {blanco2023role}
\bibfield{author}{\bibinfo{person}{Alexandre Blanco-Gonzalez}, \bibinfo{person}{Alfonso Cabezon}, \bibinfo{person}{Alejandro Seco-Gonzalez}, \bibinfo{person}{Daniel Conde-Torres}, \bibinfo{person}{Paula Antelo-Riveiro}, \bibinfo{person}{Angel Pineiro}, {and} \bibinfo{person}{Rebeca Garcia-Fandino}.} \bibinfo{year}{2023}\natexlab{}.
\newblock \showarticletitle{The role of AI in drug discovery: challenges, opportunities, and strategies}.
\newblock \bibinfo{journal}{\emph{Pharmaceuticals}} \bibinfo{volume}{16}, \bibinfo{number}{6} (\bibinfo{year}{2023}), \bibinfo{pages}{891}.
\newblock


\bibitem[Campello et~al\mbox{.}(2013)]%
        {campello2013density}
\bibfield{author}{\bibinfo{person}{Ricardo~JGB Campello}, \bibinfo{person}{Davoud Moulavi}, {and} \bibinfo{person}{J{\"o}rg Sander}.} \bibinfo{year}{2013}\natexlab{}.
\newblock \showarticletitle{Density-based clustering based on hierarchical density estimates}. In \bibinfo{booktitle}{\emph{Pacific-Asia conference on knowledge discovery and data mining}}. Springer, \bibinfo{pages}{160--172}.
\newblock


\bibitem[Choi et~al\mbox{.}(2023)]%
        {choi2023dualstreamneuralnetworkexplains}
\bibfield{author}{\bibinfo{person}{Minkyu Choi}, \bibinfo{person}{Kuan Han}, \bibinfo{person}{Xiaokai Wang}, \bibinfo{person}{Yizhen Zhang}, {and} \bibinfo{person}{Zhongming Liu}.} \bibinfo{year}{2023}\natexlab{}.
\newblock \bibinfo{title}{A Dual-Stream Neural Network Explains the Functional Segregation of Dorsal and Ventral Visual Pathways in Human Brains}.
\newblock
\showeprint[arxiv]{2310.13849}~[cs.CV]
\urldef\tempurl%
\url{https://arxiv.org/abs/2310.13849}
\showURL{%
\tempurl}


\bibitem[Dara et~al\mbox{.}(2022)]%
        {dara2022machine}
\bibfield{author}{\bibinfo{person}{Suresh Dara}, \bibinfo{person}{Swetha Dhamercherla}, \bibinfo{person}{Surender~Singh Jadav}, \bibinfo{person}{CH~Madhu Babu}, {and} \bibinfo{person}{Mohamed~Jawed Ahsan}.} \bibinfo{year}{2022}\natexlab{}.
\newblock \showarticletitle{Machine learning in drug discovery: a review}.
\newblock \bibinfo{journal}{\emph{Artificial intelligence review}} \bibinfo{volume}{55}, \bibinfo{number}{3} (\bibinfo{year}{2022}), \bibinfo{pages}{1947--1999}.
\newblock


\bibitem[Duede et~al\mbox{.}(2024)]%
        {duede2024oil}
\bibfield{author}{\bibinfo{person}{Eamon Duede}, \bibinfo{person}{William Dolan}, \bibinfo{person}{Andr{\'e} Bauer}, \bibinfo{person}{Ian Foster}, {and} \bibinfo{person}{Karim Lakhani}.} \bibinfo{year}{2024}\natexlab{}.
\newblock \showarticletitle{Oil \& water? diffusion of ai within and across scientific fields}.
\newblock \bibinfo{journal}{\emph{arXiv preprint arXiv:2405.15828}} (\bibinfo{year}{2024}).
\newblock


\bibitem[Gao and Wang(2023)]%
        {gao2023quantifying}
\bibfield{author}{\bibinfo{person}{Jian Gao} {and} \bibinfo{person}{Dashun Wang}.} \bibinfo{year}{2023}\natexlab{}.
\newblock \showarticletitle{Quantifying the benefit of artificial intelligence for scientific research}.
\newblock \bibinfo{journal}{\emph{arXiv preprint arXiv:2304.10578}} (\bibinfo{year}{2023}).
\newblock


\bibitem[Grover and Leskovec(2016)]%
        {grover2016node2vec}
\bibfield{author}{\bibinfo{person}{Aditya Grover} {and} \bibinfo{person}{Jure Leskovec}.} \bibinfo{year}{2016}\natexlab{}.
\newblock \showarticletitle{node2vec: Scalable feature learning for networks}. In \bibinfo{booktitle}{\emph{Proceedings of the 22nd ACM SIGKDD international conference on Knowledge discovery and data mining}}. \bibinfo{pages}{855--864}.
\newblock


\bibitem[Guo et~al\mbox{.}(2021)]%
        {guo2021artificial}
\bibfield{author}{\bibinfo{person}{Kai Guo}, \bibinfo{person}{Zhenze Yang}, \bibinfo{person}{Chi-Hua Yu}, {and} \bibinfo{person}{Markus~J Buehler}.} \bibinfo{year}{2021}\natexlab{}.
\newblock \showarticletitle{Artificial intelligence and machine learning in design of mechanical materials}.
\newblock \bibinfo{journal}{\emph{Materials Horizons}} \bibinfo{volume}{8}, \bibinfo{number}{4} (\bibinfo{year}{2021}), \bibinfo{pages}{1153--1172}.
\newblock


\bibitem[Jumper et~al\mbox{.}(2021)]%
        {jumper2021highly}
\bibfield{author}{\bibinfo{person}{John Jumper}, \bibinfo{person}{Richard Evans}, \bibinfo{person}{Alexander Pritzel}, \bibinfo{person}{Tim Green}, \bibinfo{person}{Michael Figurnov}, \bibinfo{person}{Olaf Ronneberger}, \bibinfo{person}{Kathryn Tunyasuvunakool}, \bibinfo{person}{Russ Bates}, \bibinfo{person}{Augustin {\v{Z}}{\'\i}dek}, \bibinfo{person}{Anna Potapenko}, {et~al\mbox{.}}} \bibinfo{year}{2021}\natexlab{}.
\newblock \showarticletitle{Highly accurate protein structure prediction with AlphaFold}.
\newblock \bibinfo{journal}{\emph{nature}} \bibinfo{volume}{596}, \bibinfo{number}{7873} (\bibinfo{year}{2021}), \bibinfo{pages}{583--589}.
\newblock


\bibitem[Katz(1953)]%
        {katz1953new}
\bibfield{author}{\bibinfo{person}{Leo Katz}.} \bibinfo{year}{1953}\natexlab{}.
\newblock \showarticletitle{A new status index derived from sociometric analysis}.
\newblock \bibinfo{journal}{\emph{Psychometrika}} \bibinfo{volume}{18}, \bibinfo{number}{1} (\bibinfo{year}{1953}), \bibinfo{pages}{39--43}.
\newblock


\bibitem[Kuenneth and Ramprasad(2023)]%
        {kuenneth2023polybert}
\bibfield{author}{\bibinfo{person}{C. Kuenneth} {and} \bibinfo{person}{R. Ramprasad}.} \bibinfo{year}{2023}\natexlab{}.
\newblock \showarticletitle{polyBERT: a chemical language model to enable fully machine-driven ultrafast polymer informatics}.
\newblock \bibinfo{journal}{\emph{Nature Communications}}  \bibinfo{volume}{14} (\bibinfo{year}{2023}), \bibinfo{pages}{4099}.
\newblock
\href{https://doi.org/10.1038/s41467-023-39868-6}{doi:\nolinkurl{10.1038/s41467-023-39868-6}}


\bibitem[Leontidis(2024)]%
        {leontidis2024science}
\bibfield{author}{\bibinfo{person}{Georgios Leontidis}.} \bibinfo{year}{2024}\natexlab{}.
\newblock \showarticletitle{Science in the age of AI: How artificial intelligence is changing the nature and method of scientific research}.
\newblock  (\bibinfo{year}{2024}).
\newblock


\bibitem[Lewis et~al\mbox{.}(2020)]%
        {lewis2020retrieval}
\bibfield{author}{\bibinfo{person}{Patrick Lewis}, \bibinfo{person}{Ethan Perez}, \bibinfo{person}{Aleksandra Piktus}, \bibinfo{person}{Fabio Petroni}, \bibinfo{person}{Vladimir Karpukhin}, \bibinfo{person}{Naman Goyal}, \bibinfo{person}{Heinrich K{\"u}ttler}, \bibinfo{person}{Mike Lewis}, \bibinfo{person}{Wen-tau Yih}, \bibinfo{person}{Tim Rockt{\"a}schel}, {et~al\mbox{.}}} \bibinfo{year}{2020}\natexlab{}.
\newblock \showarticletitle{Retrieval-augmented generation for knowledge-intensive nlp tasks}.
\newblock \bibinfo{journal}{\emph{Advances in Neural Information Processing Systems}}  \bibinfo{volume}{33} (\bibinfo{year}{2020}), \bibinfo{pages}{9459--9474}.
\newblock


\bibitem[Li et~al\mbox{.}(2020)]%
        {li2020ai}
\bibfield{author}{\bibinfo{person}{Jiali Li}, \bibinfo{person}{Kaizhuo Lim}, \bibinfo{person}{Haitao Yang}, \bibinfo{person}{Zekun Ren}, \bibinfo{person}{Shreyaa Raghavan}, \bibinfo{person}{Po-Yen Chen}, \bibinfo{person}{Tonio Buonassisi}, {and} \bibinfo{person}{Xiaonan Wang}.} \bibinfo{year}{2020}\natexlab{}.
\newblock \showarticletitle{AI applications through the whole life cycle of material discovery}.
\newblock \bibinfo{journal}{\emph{Matter}} \bibinfo{volume}{3}, \bibinfo{number}{2} (\bibinfo{year}{2020}), \bibinfo{pages}{393--432}.
\newblock


\bibitem[Li et~al\mbox{.}(2023)]%
        {doi:10.1126/sciadv.adh3784}
\bibfield{author}{\bibinfo{person}{Junwei Li}, \bibinfo{person}{Yanhao Lin}, \bibinfo{person}{Thomas Meier}, \bibinfo{person}{Zhipan Liu}, \bibinfo{person}{Wei Yang}, \bibinfo{person}{Ho kwang Mao}, \bibinfo{person}{Shengcai Zhu}, {and} \bibinfo{person}{Qingyang Hu}.} \bibinfo{year}{2023}\natexlab{}.
\newblock \showarticletitle{Silica-water superstructure and one-dimensional superionic conduit in Earth’s mantle}.
\newblock \bibinfo{journal}{\emph{Science Advances}} \bibinfo{volume}{9}, \bibinfo{number}{35} (\bibinfo{year}{2023}), \bibinfo{pages}{eadh3784}.
\newblock
\href{https://doi.org/10.1126/sciadv.adh3784}{doi:\nolinkurl{10.1126/sciadv.adh3784}}
\showeprint{https://www.science.org/doi/pdf/10.1126/sciadv.adh3784}


\bibitem[Lin(2004)]%
        {lin2004rouge}
\bibfield{author}{\bibinfo{person}{Chin-Yew Lin}.} \bibinfo{year}{2004}\natexlab{}.
\newblock \showarticletitle{Rouge: A package for automatic evaluation of summaries}. In \bibinfo{booktitle}{\emph{Text summarization branches out}}. \bibinfo{pages}{74--81}.
\newblock


\bibitem[Mak et~al\mbox{.}(2023)]%
        {mak2023artificial}
\bibfield{author}{\bibinfo{person}{Kit-Kay Mak}, \bibinfo{person}{Yi-Hang Wong}, {and} \bibinfo{person}{Mallikarjuna~Rao Pichika}.} \bibinfo{year}{2023}\natexlab{}.
\newblock \showarticletitle{Artificial intelligence in drug discovery and development}.
\newblock \bibinfo{journal}{\emph{Drug Discovery and Evaluation: Safety and Pharmacokinetic Assays}} (\bibinfo{year}{2023}), \bibinfo{pages}{1--38}.
\newblock


\bibitem[Mei et~al\mbox{.}(2007)]%
        {mei2007automatic}
\bibfield{author}{\bibinfo{person}{Qiaozhu Mei}, \bibinfo{person}{Xuehua Shen}, {and} \bibinfo{person}{ChengXiang Zhai}.} \bibinfo{year}{2007}\natexlab{}.
\newblock \showarticletitle{Automatic labeling of multinomial topic models}. In \bibinfo{booktitle}{\emph{Proceedings of the 13th ACM SIGKDD international conference on Knowledge discovery and data mining}}. \bibinfo{pages}{490--499}.
\newblock


\bibitem[Meller et~al\mbox{.}(2023)]%
        {meller2023predicting}
\bibfield{author}{\bibinfo{person}{A. Meller}, \bibinfo{person}{M. Ward}, \bibinfo{person}{J. Borowsky}, \bibinfo{person}{M. Kshirsagar}, \bibinfo{person}{J.~M. Lotthammer}, \bibinfo{person}{F. Oviedo}, \bibinfo{person}{J.~L. Ferres}, {and} \bibinfo{person}{G.~R. Bowman}.} \bibinfo{year}{2023}\natexlab{}.
\newblock \showarticletitle{Predicting locations of cryptic pockets from single protein structures using the PocketMiner graph neural network}.
\newblock \bibinfo{journal}{\emph{Nature Communications}} \bibinfo{volume}{14}, \bibinfo{number}{1} (\bibinfo{date}{Mar} \bibinfo{year}{2023}), \bibinfo{pages}{1177}.
\newblock
\href{https://doi.org/10.1038/s41467-023-36699-3}{doi:\nolinkurl{10.1038/s41467-023-36699-3}}
\newblock
\shownote{PMID: 36859488; PMCID: PMC9977097}.


\bibitem[{\"O}zer et~al\mbox{.}(2024)]%
        {ozer2024link}
\bibfield{author}{\bibinfo{person}{{\c{S}}{\"u}kr{\"u} Demir~{\.I}nan {\"O}zer}, \bibinfo{person}{G{\"u}nce~Keziban Orman}, {and} \bibinfo{person}{Vincent Labatut}.} \bibinfo{year}{2024}\natexlab{}.
\newblock \showarticletitle{Link Prediction in Bipartite Networks}.
\newblock \bibinfo{journal}{\emph{arXiv preprint arXiv:2406.06658}} (\bibinfo{year}{2024}).
\newblock


\bibitem[Perc(2010)]%
        {perc2010growth}
\bibfield{author}{\bibinfo{person}{Matja{\v{z}} Perc}.} \bibinfo{year}{2010}\natexlab{}.
\newblock \showarticletitle{Growth and structure of Slovenia’s scientific collaboration network}.
\newblock \bibinfo{journal}{\emph{Journal of Informetrics}} \bibinfo{volume}{4}, \bibinfo{number}{4} (\bibinfo{year}{2010}), \bibinfo{pages}{475--482}.
\newblock


\bibitem[Perozzi et~al\mbox{.}(2014)]%
        {perozzi2014deepwalk}
\bibfield{author}{\bibinfo{person}{Bryan Perozzi}, \bibinfo{person}{Rami Al-Rfou}, {and} \bibinfo{person}{Steven Skiena}.} \bibinfo{year}{2014}\natexlab{}.
\newblock \showarticletitle{Deepwalk: Online learning of social representations}. In \bibinfo{booktitle}{\emph{Proceedings of the 20th ACM SIGKDD international conference on Knowledge discovery and data mining}}. \bibinfo{pages}{701--710}.
\newblock


\bibitem[Qureshi et~al\mbox{.}(2023)]%
        {qureshi2023ai}
\bibfield{author}{\bibinfo{person}{Rizwan Qureshi}, \bibinfo{person}{Muhammad Irfan}, \bibinfo{person}{Taimoor~Muzaffar Gondal}, \bibinfo{person}{Sheheryar Khan}, \bibinfo{person}{Jia Wu}, \bibinfo{person}{Muhammad~Usman Hadi}, \bibinfo{person}{John Heymach}, \bibinfo{person}{Xiuning Le}, \bibinfo{person}{Hong Yan}, {and} \bibinfo{person}{Tanvir Alam}.} \bibinfo{year}{2023}\natexlab{}.
\newblock \showarticletitle{AI in drug discovery and its clinical relevance}.
\newblock \bibinfo{journal}{\emph{Heliyon}} \bibinfo{volume}{9}, \bibinfo{number}{7} (\bibinfo{year}{2023}).
\newblock


\bibitem[Sellam et~al\mbox{.}(2020)]%
        {sellam2020bleurt}
\bibfield{author}{\bibinfo{person}{Thibault Sellam}, \bibinfo{person}{Dipanjan Das}, {and} \bibinfo{person}{Ankur~P Parikh}.} \bibinfo{year}{2020}\natexlab{}.
\newblock \showarticletitle{BLEURT: Learning robust metrics for text generation}.
\newblock \bibinfo{journal}{\emph{arXiv preprint arXiv:2004.04696}} (\bibinfo{year}{2020}).
\newblock


\bibitem[Sha et~al\mbox{.}(2020)]%
        {sha2020artificial}
\bibfield{author}{\bibinfo{person}{Wuxin Sha}, \bibinfo{person}{Yaqing Guo}, \bibinfo{person}{Qing Yuan}, \bibinfo{person}{Shun Tang}, \bibinfo{person}{Xinfang Zhang}, \bibinfo{person}{Songfeng Lu}, \bibinfo{person}{Xin Guo}, \bibinfo{person}{Yuan-Cheng Cao}, {and} \bibinfo{person}{Shijie Cheng}.} \bibinfo{year}{2020}\natexlab{}.
\newblock \showarticletitle{Artificial intelligence to power the future of materials science and engineering}.
\newblock \bibinfo{journal}{\emph{Advanced Intelligent Systems}} \bibinfo{volume}{2}, \bibinfo{number}{4} (\bibinfo{year}{2020}), \bibinfo{pages}{1900143}.
\newblock


\bibitem[Si et~al\mbox{.}(2024)]%
        {si2024can}
\bibfield{author}{\bibinfo{person}{Chenglei Si}, \bibinfo{person}{Diyi Yang}, {and} \bibinfo{person}{Tatsunori Hashimoto}.} \bibinfo{year}{2024}\natexlab{}.
\newblock \showarticletitle{Can LLMs Generate Novel Research Ideas?}
\newblock \bibinfo{journal}{\emph{arXiv preprint arXiv:2409.04109}} (\bibinfo{year}{2024}).
\newblock


\bibitem[Su et~al\mbox{.}(2022)]%
        {su2022one}
\bibfield{author}{\bibinfo{person}{Hongjin Su}, \bibinfo{person}{Weijia Shi}, \bibinfo{person}{Jungo Kasai}, \bibinfo{person}{Yizhong Wang}, \bibinfo{person}{Yushi Hu}, \bibinfo{person}{Mari Ostendorf}, \bibinfo{person}{Wen-tau Yih}, \bibinfo{person}{Noah~A Smith}, \bibinfo{person}{Luke Zettlemoyer}, {and} \bibinfo{person}{Tao Yu}.} \bibinfo{year}{2022}\natexlab{}.
\newblock \showarticletitle{One embedder, any task: Instruction-finetuned text embeddings}.
\newblock \bibinfo{journal}{\emph{arXiv preprint arXiv:2212.09741}} (\bibinfo{year}{2022}).
\newblock


\bibitem[Sun et~al\mbox{.}(2014)]%
        {sun2014analyzing}
\bibfield{author}{\bibinfo{person}{Huan Sun}, \bibinfo{person}{Mudhakar Srivatsa}, \bibinfo{person}{Shulong Tan}, \bibinfo{person}{Yang Li}, \bibinfo{person}{Lance~M Kaplan}, \bibinfo{person}{Shu Tao}, {and} \bibinfo{person}{Xifeng Yan}.} \bibinfo{year}{2014}\natexlab{}.
\newblock \showarticletitle{Analyzing expert behaviors in collaborative networks}. In \bibinfo{booktitle}{\emph{Proceedings of the 20th ACM SIGKDD international conference on Knowledge discovery and data mining}}. \bibinfo{pages}{1486--1495}.
\newblock


\bibitem[Tang et~al\mbox{.}(2016)]%
        {tang2016visualizing}
\bibfield{author}{\bibinfo{person}{Jian Tang}, \bibinfo{person}{Jingzhou Liu}, \bibinfo{person}{Ming Zhang}, {and} \bibinfo{person}{Qiaozhu Mei}.} \bibinfo{year}{2016}\natexlab{}.
\newblock \showarticletitle{Visualizing large-scale and high-dimensional data}. In \bibinfo{booktitle}{\emph{Proceedings of the 25th international conference on world wide web}}. \bibinfo{pages}{287--297}.
\newblock


\bibitem[Tang et~al\mbox{.}(2015)]%
        {tang2015line}
\bibfield{author}{\bibinfo{person}{Jian Tang}, \bibinfo{person}{Meng Qu}, \bibinfo{person}{Mingzhe Wang}, \bibinfo{person}{Ming Zhang}, \bibinfo{person}{Jun Yan}, {and} \bibinfo{person}{Qiaozhu Mei}.} \bibinfo{year}{2015}\natexlab{}.
\newblock \showarticletitle{Line: Large-scale information network embedding}. In \bibinfo{booktitle}{\emph{Proceedings of the 24th international conference on world wide web}}. \bibinfo{pages}{1067--1077}.
\newblock


\bibitem[Vasques~Filho and O'Neale(2018)]%
        {vasques2018degree}
\bibfield{author}{\bibinfo{person}{Demival Vasques~Filho} {and} \bibinfo{person}{Dion~RJ O'Neale}.} \bibinfo{year}{2018}\natexlab{}.
\newblock \showarticletitle{Degree distributions of bipartite networks and their projections}.
\newblock \bibinfo{journal}{\emph{Physical Review E}} \bibinfo{volume}{98}, \bibinfo{number}{2} (\bibinfo{year}{2018}), \bibinfo{pages}{022307}.
\newblock


\bibitem[Wang et~al\mbox{.}(2023)]%
        {wang2023scientific}
\bibfield{author}{\bibinfo{person}{Hanchen Wang}, \bibinfo{person}{Tianfan Fu}, \bibinfo{person}{Yuanqi Du}, \bibinfo{person}{Wenhao Gao}, \bibinfo{person}{Kexin Huang}, \bibinfo{person}{Ziming Liu}, \bibinfo{person}{Payal Chandak}, \bibinfo{person}{Shengchao Liu}, \bibinfo{person}{Peter Van~Katwyk}, \bibinfo{person}{Andreea Deac}, {et~al\mbox{.}}} \bibinfo{year}{2023}\natexlab{}.
\newblock \showarticletitle{Scientific discovery in the age of artificial intelligence}.
\newblock \bibinfo{journal}{\emph{Nature}} \bibinfo{volume}{620}, \bibinfo{number}{7972} (\bibinfo{year}{2023}), \bibinfo{pages}{47--60}.
\newblock


\bibitem[Wang et~al\mbox{.}(2014)]%
        {wang2014discover}
\bibfield{author}{\bibinfo{person}{Yan Wang}, \bibinfo{person}{Jie Liang}, {and} \bibinfo{person}{Jianguo Lu}.} \bibinfo{year}{2014}\natexlab{}.
\newblock \showarticletitle{Discover hidden web properties by random walk on bipartite graph}.
\newblock \bibinfo{journal}{\emph{Information retrieval}}  \bibinfo{volume}{17} (\bibinfo{year}{2014}), \bibinfo{pages}{203--228}.
\newblock


\bibitem[Xu et~al\mbox{.}(2024)]%
        {xu2024ai}
\bibfield{author}{\bibinfo{person}{Ruoxi Xu}, \bibinfo{person}{Yingfei Sun}, \bibinfo{person}{Mengjie Ren}, \bibinfo{person}{Shiguang Guo}, \bibinfo{person}{Ruotong Pan}, \bibinfo{person}{Hongyu Lin}, \bibinfo{person}{Le Sun}, {and} \bibinfo{person}{Xianpei Han}.} \bibinfo{year}{2024}\natexlab{}.
\newblock \showarticletitle{AI for social science and social science of AI: A survey}.
\newblock \bibinfo{journal}{\emph{Information Processing \& Management}} \bibinfo{volume}{61}, \bibinfo{number}{3} (\bibinfo{year}{2024}), \bibinfo{pages}{103665}.
\newblock


\bibitem[Yang et~al\mbox{.}(2015)]%
        {yang2015evaluating}
\bibfield{author}{\bibinfo{person}{Yang Yang}, \bibinfo{person}{Ryan~N Lichtenwalter}, {and} \bibinfo{person}{Nitesh~V Chawla}.} \bibinfo{year}{2015}\natexlab{}.
\newblock \showarticletitle{Evaluating link prediction methods}.
\newblock \bibinfo{journal}{\emph{Knowledge and Information Systems}}  \bibinfo{volume}{45} (\bibinfo{year}{2015}), \bibinfo{pages}{751--782}.
\newblock


\bibitem[Zhang et~al\mbox{.}(2024)]%
        {zhang2024massw}
\bibfield{author}{\bibinfo{person}{Xingjian Zhang}, \bibinfo{person}{Yutong Xie}, \bibinfo{person}{Jin Huang}, \bibinfo{person}{Jinge Ma}, \bibinfo{person}{Zhaoying Pan}, \bibinfo{person}{Qijia Liu}, \bibinfo{person}{Ziyang Xiong}, \bibinfo{person}{Tolga Ergen}, \bibinfo{person}{Dongsub Shim}, \bibinfo{person}{Honglak Lee}, {et~al\mbox{.}}} \bibinfo{year}{2024}\natexlab{}.
\newblock \showarticletitle{MASSW: A New Dataset and Benchmark Tasks for AI-Assisted Scientific Workflows}.
\newblock \bibinfo{journal}{\emph{arXiv preprint arXiv:2406.06357}} (\bibinfo{year}{2024}).
\newblock


\end{thebibliography}

\newpage
\appendix
\section{AI4Science Dataset}

\subsection{Curation of Literature}
\label{app:data-curation}

\paragraph{Data source. }

\begin{itemize} 
    \item For the science domain, we obtained our data via the NCBI Entrez database, encompassing the following papers: \textit{Nature, Science, PNAS, Nature Communications, Science Advances}. Only papers with both title and abstract available and within 2014 to 2024 are kept. A brief statistics of science domain papers are as follows: 
    \begin{table}[h]
    \centering
    \begin{tabular}{l r r}
        \toprule
        \textbf{Journal} & \textbf{\#Pub.} & \textbf{\#AI4Science Pub.} \\
        \midrule
        Nature & 10,890 & 254\\ 
        Science &  10,998 & 207\\ 
        PNAS & 35,556 & 997\\ 
        Nature Communications & 54,243 & 2,027 \\ 
        Science Advances &  12,086 & 488\\ 
        \hline
        Total & 123,773 & 3,973 \\
        \bottomrule
    \end{tabular}
    \caption{Science Domain dataset statistics}
    \vspace{-25pt}
    \label{tab:science_statistics} 
    \end{table}
    \item For AI domain, we obtained our data from MASSW \cite{zhang2024massw}, which encompasses the following seven top AI conferences: \textit{AAAI, IJCAI, ICLR, ICML, NeurIPS, WWW, SIGKDD}. Only papers with both title and abstract available and within 2014 to 2024 are kept. A brief statistics of AI domain papers are as follows: 
    \begin{table}[h]
    \centering
    \begin{tabular}{l r r}
        \toprule
        \textbf{Conference} & \textbf{\#Pub.} & \textbf{\#AI4Science Pub.} \\
        \midrule
        AAAI &  8,994 & 915\\ 
        IJCAI & 2,953 & 271\\ 
        ICLR  &  2,907 & 205\\
        ICML & 5,641 & 530\\ 
        NeurIPS & 11,250 & 889\\ 
        WWW & 3,777 & 241\\
        SIGKDD &   3,251    & 488 \\
        \hline
        Total & 38,773 & 3,539 \\
        \bottomrule
    \end{tabular}
    \caption{AI Domain dataset statistics
    }
    \vspace{-25pt}
    \label{tab:science_statistics} 
    \end{table}

\end{itemize}

\subsection{Extraction with LLMs}
\label{app:data-extract}

\paragraph{Extracting key aspects}
As aforementioned, we asked large language models to extract the following three key aspects: \textbf{Scientific Problem}, \textbf{AI Method}, \textbf{Usage}. To help LLM better understand the three aspects, we further break down the three key aspects into the following six aspects when prompting the model: 

\begin{itemize}
    \item Scientific Problem (keyword/keyphrase): A few keywords or key phrases that summarize the primary scientific problem addressed in the paper.
    \item Scientific Problem (definition): A detailed definition of the primary scientific problem addressed in the paper.
    \item Scientific Problem discipline: The scientific discipline that the primary scientific problem addressed in the paper falls into.
    \item AI method (keywords/keyphrase): A few keywords or key phrases that summarize the AI method used in the paper.
    \item AI method (definition): A detailed definition of the AI method used in the paper.
    \item AI Usage: A detailed explanation of how the AI method
is specifically applied to the scientific problem.

\end{itemize} 
\paragraph{Prompting Template}

We used the following prompt to extract the six key aspects of a paper: 

\begin{framed}
\noindent \#\# Background and Task Description \newline
You are an expert in both science and artificial intelligence (AI), where AI generally refers to intelligence exhibited by machines (particularly computer systems), including models, algorithms, etc. Given a scientific paper, your task is to extract the following aspects from the title and abstract:  \newline \newline
**Problem (keyword/keyphrase):** A keyword or a keyphrase that summarizes the main problem to be addressed in this paper. \newline
**Problem (definition):** The detailed definition of the problem. \newline
**Problem discipline:** The discipline in which the main problem best fits. \newline
**Method (keyword/keyphrase):** A keyword or a keyphrase that summarizes the main method used in this paper to address the above problem. \newline
**Method (definition):** The detailed definition of the method. \newline
**Usage:** How the method is specifically applied to address the problem. \newline \newline
\#\# Scientific Paper to Be Extracted  \newline
Title: \{title\}  \newline
Abstract: \{abstract\} \newline \newline
\#\# Requirements \newline 
* Please do not include any method-specific information in problem extraction. \newline
* Similarly, please do not include any problem-specific information in method extraction. The extracted method description (keyword/keyphrase and definition) should be generic and can be applied across all application domains.  \newline
* Please do not use abbreviations as keywords/keyphrases. \newline \newline
Please output the extraction results in the JSON format as below. Fields could be "N/A" if no relevant information can be found in the paper title and abstract. \newline
\{ \newline
    "Problem (keyword/keyphrase)": "...", \newline
    "Problem (definition)": "...", \newline
    "Problem Discipline": "...", \newline
    "Method (keyword/keyphrase)": "...", \newline
    "Method (definition)": "...", \newline
    "Usage": "..." \newline
\}
\end{framed}

\paragraph{Classifying AI4Science}

Given the extracted key aspects of papers, we leveraged the following prompt to ask LLM to judge whether the paper solves scientific problems and whether the paper uses AI methods: 

\begin{framed}
\noindent \#\# Task Description \newline
Given the extraction results of a research paper, please determine if the main research problem is a scientific problem from traditional disciplines in Science (not including disciplines like Computer Science and Information Science), and if the main method involves the use of Artificial Intelligence. \newline \newline
\#\# Research Paper to Be Extracted \newline 
Title: \{title\} \newline
Abstract: \{abstract\} \newline \newline
\#\# Extraction Results \newline
\{ \newline
    results \newline
\} \newline
\#\# Note \newline 
* Don't consider research problem disciplines that involves Computer Science, Information/Data Science as traditional scientific problems. \newline \newline
\#\# Response Format: \newline
Only a dictionary containing the following \newline
\{ \newline
    "Scientific problem": True/False, \newline
    "AI method": True/False, \newline
\} \newline
\end{framed}

\paragraph{Model.}
For the LLM extraction, we utilized \texttt{gpt-4o-mini-2024-07-18}, which represents a good balance between extraction capability and affordability. We use default generation hyperparameters (e.g., temperature) for all extraction tasks.

\paragraph{Human evaluation. }


To evaluate LLM's ability as an annotator for scholarly papers, we conducted a human evaluation of 50 papers from scientific journals papers and 50 papers from AI conferences, which were randomly selected within our dataset.

Two trained human experts who are familiar with reading scientific literature are assigned to annotate key aspects extraction of each paper, based on the title and abstract, following a carefully designed codebook. They are asked to judge whether the LLM extraction of key aspects is accurate. After that, they are asked whether the paper of interest is an AI4Science paper. 


The results are shown in Table \ref{tab:human-eval-accuracy} \ref{tab:human-eval-judge}, in which the human agreement is calculated by treating one annotation as the reference and the other as the prediction for each paper. The average extraction accuracy reported by human annotators is 0.910. Together with high human agreement F1 scores (0.876 and 0.946 respectively), this shows that LLM extractions of key aspects are aligned with human preferences. The evaluations for AI4Science classification are also reported, where the F1 score reaches 0.836 for publications from scientific journals. For publications from AI conferences, the F1 score is relatively low (0.647), because of the difficulty in distinguishing research problems in Computer Science domains and general Science domains. However, the high Recall score (0.866) provides a guarantee of avoiding missing important AI4Science work. The difficulty of AI4Science classification is further illustrated by relatively low human agreement(F1 score of 0.632 for both genres), which shows the lack of universally accepted standard of AI4Science. Moreover, this shows that our LLM extraction results are more consistent compared with a group of human annotators.

The high extraction accuracy indicated by our human evaluation also aligns with human evaluation results from recent studies \cite{zhang2024massw}, where LLMs have been used to extract scientific workflows from titles and abstracts.

 \begin{table}[h]
    \centering
    
    \begin{tabular}{lcc}
        \toprule
          & \textbf{Extraction acc.} & \textbf{Human agreement F1} \\
        \midrule
        Scientific pubs. & 0.890 & 0.876  \\
        AI pubs. & 0.930 & 0.946  \\
        \bottomrule
    \end{tabular}
    \caption{Human evaluation of \texttt{gpt-4o-mini-2024-07-18} extractions. }
    \vspace{-25pt}
    \label{tab:human-eval-accuracy} 
    \end{table}

 \begin{table}[h]
    \centering
    
    \begin{tabular}{lcccc}
        \toprule
        & \multicolumn{3}{c}{\textbf{Classification acc.}} & \textbf{Human Agreement F1}\\
        
        &  Prec. & Recall & F1 & \\
        \midrule
        Scientific pubs. & 0.833 & 0.840 & 0.836 & 0.632\\
        AI pubs. & 0.516 & 0.866 & 0.647 & 0.632 \\
        \bottomrule
    \end{tabular}
    \caption{Human evaluation of \texttt{gpt-4o-mini-2024-07-18} AI4Science Classification. }
    \vspace{-25pt}
    \label{tab:human-eval-judge} 
    \end{table}

\paragraph{Extraction examples. }

Listed below are some examples of key aspects extracted by LLM within the human annotation set and whether the paper involves the use of AI methods or tries to solve scientific problems.

Scientific journal paper examples: Please refer to table \ref{tab:sci-examples}

\begin{table*}[htbp]
\begin{tabular}{|p{16cm}|}
\hline
\textbf{Title:} A machine learning approach to integrate big data for precision medicine in acute myeloid leukemia \\ 
\textbf{Abstract:} Cancers that appear pathologically similar often respond differently to the same drug regimens. Methods to better match patients to drugs are in high demand. We demonstrate a promising approach to identify robust molecular markers for targeted treatment of acute myeloid leukemia (AML) by introducing: data from 30 AML patients including genome-wide gene expression profiles and in vitro sensitivity to 160 chemotherapy drugs, a computational method to identify reliable gene expression markers for drug sensitivity by incorporating multi-omic prior information relevant to each gene's potential to drive cancer. We show that our method outperforms several state-of-the-art approaches in identifying molecular markers replicated in validation data and predicting drug sensitivity accurately. Finally, we identify SMARCA4 as a marker and driver of sensitivity to topoisomerase II inhibitors, mitoxantrone, and etoposide, in AML by showing that cell lines transduced to have high SMARCA4 expression reveal dramatically increased sensitivity to these agents. \\ 
\textbf{LLM Summarized Core Aspects:} \\ 
\textbf{Problem (keyword/keyphrase):} Patient-Drug Matching in Acute Myeloid Leukemia \\ 
\textbf{Problem (definition):} The need for identifying effective drug regimens for patients with acute myeloid leukemia, as similar cancers can respond differently to the same treatments, necessitating a method to better match patients with suitable therapies based on molecular markers. \\ 
\textbf{Problem discipline:} Medical Science/Oncology \\ 
\textbf{Method (keyword/keyphrase):} Computational Method for Gene Expression Analysis \\ 
\textbf{Method (definition):} A computational approach that utilizes data analysis techniques to identify reliable markers from gene expression profiles, potentially incorporating additional omic data to enhance the predictive accuracy for drug sensitivity. \\ 
\textbf{Usage:} The method is applied by analyzing gene expression data from acute myeloid leukemia patients to identify molecular markers that are associated with sensitivity to various chemotherapy drugs, thereby improving the precision of treatment matching for individual patients. \\ 
\textbf{AI4Science Classification:} \\ 
\textbf{Scientific problem:} True \\ 
\textbf{AI method:} True \\ 
\hline
\textbf{Title:} Dynamic microscale flow patterning using electrical modulation of zeta potential \\ 
\textbf{Abstract:} Cancers that appear pathologically similar often respond differently to the same drug regimens. Methods to better match patients to drugs are in high demand. We demonstrate a promising approach to identify robust molecular markers for targeted treatment of acute myeloid leukemia (AML) by introducing: data from 30 AML patients including genome-wide gene expression profiles and in vitro sensitivity to 160 chemotherapy drugs, a computational method to identify reliable gene expression markers for drug sensitivity by incorporating multi-omic prior information relevant to each gene's potential to drive cancer. We show that our method outperforms several state-of-the-art approaches in identifying molecular markers replicated in validation data and predicting drug sensitivity accurately. Finally, we identify SMARCA4 as a marker and driver of sensitivity to topoisomerase II inhibitors, mitoxantrone, and etoposide, in AML by showing that cell lines transduced to have high SMARCA4 expression reveal dramatically increased sensitivity to these agents. \\ 
\textbf{LLM Summarized Core Aspects:} \\ 
\textbf{Problem (keyword/keyphrase):} Microscale flow control \\
\textbf{Problem (definition):} The limitations associated with moving fluids at the microscale, particularly the constraints imposed by discrete channels, mechanical valves, and fixed geometries, which restrict the ability to manipulate fluid dynamics effectively. \\
\textbf{Problem discipline:} Fluid Mechanics \\ 
\textbf{Method (keyword/keyphrase):} Field-effect electroosmosis \\
\textbf{Method (definition):} A technique that utilizes electric fields to induce fluid flow in a medium, allowing for the manipulation of fluid dynamics through the controlled application of electrical signals. \\ 
\textbf{Usage:} The method is applied by controlling gate electrodes with an alternating current voltage to generate dynamic flow patterns in a fluidic chamber, enabling real-time modulation of flow characteristics without physical barriers. \\ 
\textbf{AI4Science Classification:} \\ 
\textbf{Scientific problem:} True \\ 
\textbf{AI method:} False \\
\hline
\end{tabular}
\caption{Examples of scientific paper extraction and AI4Science judging results}
\label{tab:sci-examples}
\end{table*}

AI conference paper examples: Please refer to table \ref{tab:ai-examples}

\begin{table*}[htbp]
\begin{tabular}{|p{16cm}|}
\hline
\textbf{Title}: A region-based model for estimating urban air pollution \\
\textbf{Abstract}: Air pollution has a direct impact to human health, and data-driven air quality models are useful for evaluating population exposure to air pollutants. In this paper, we propose a novel region-based Gaussian process model for estimating urban air pollution dispersion, and applied it to a large dataset of ultrafine particle (UFP) measurements collected from a network of sensors located on several trams in the city of Zurich. We show that compared to existing grid-based models, the region-based model produces better predictions across aggregates of all time scales. The new model is appropriate for many useful user applications such as exposure assessment and anomaly detection. \\
\textbf{LLM Summarized Core Aspects:} \\ 
\textbf{Problem (keyword/keyphrase):} urban air pollution 
\textbf{Problem (definition):} The challenge of estimating and predicting the dispersion of air pollutants in urban environments, which directly impacts human health and exposure assessments.\\ \textbf{Problem discipline:} Environmental Science \\
\textbf{Method (keyword/keyphrase)}: Gaussian process model \\
\textbf{Method (definition):} A statistical method used for regression and classification tasks that relies on the principles of Bayesian inference and provides a flexible way to model complex relationships in data. \\
\textbf{Usage:} The method is applied to a large dataset of ultrafine particle measurements to estimate urban air pollution dispersion and improve prediction accuracy compared to traditional grid-based models.\\
\textbf{AI4Science Classification:} \\ 
\textbf{Scientific problem:} True \\ 
\textbf{AI method:} True \\
\hline
\textbf{Title:} Dependent Relational Gamma Process Models for Longitudinal Networks \\
\textbf{Abstract:} A probabilistic framework based on the covariate-dependent relational gamma process is developed to analyze relational data arising from longitudinal networks. The proposed framework characterizes networked nodes by nonnegative node-group memberships, which allow each node to belong to multiple latent groups simultaneously, and encodes edge probabilities between each pair of nodes using a Bernoulli Poisson link to the embedded latent space. Within the latent space, our framework models the birth and death dynamics of individual groups via a thinning function. Our framework also captures the evolution of individual node-group memberships over time using gamma Markov processes. Exploiting the recent advances in data augmentation and marginalization techniques, a simple and efficient Gibbs sampler is proposed for posterior computation. Experimental results on a simulation study and three real-world temporal network data sets demonstrate the model's capability, competitive performance and scalability compared to state-of-the-art methods. \\
\textbf{LLM Summarized Core Aspects:} \\ 
\textbf{Problem (keyword/keyphrase):} Longitudinal networks analysis \\
\textbf{Problem (definition):} The challenge of analyzing relational data that evolves over time within networks, specifically focusing on the dynamics of node-group memberships and the interactions between nodes in a temporal context \\ 
\textbf{Problem discipline:} Statistics and Network Science \\
\textbf{Method (keyword/keyphrase)}: Bayesian inference using Gibbs sampling \\
\textbf{Method (definition):} A statistical method that involves making inferences about unknown parameters in a model through the use of a sampling technique that generates samples from the posterior distribution of those parameters. \\
\textbf{Usage:} The method is applied to perform posterior computation for the parameters of the covariate-dependent relational gamma process model, enabling the analysis of the evolving structures of longitudinal networks. \\
\textbf{AI4Science Classification:} \\ 
\textbf{Scientific problem:} False \\ 
\textbf{AI method:} True \\
\hline
\end{tabular}
\caption{Examples of AI paper LLM extraction and AI4Science judging results}
\label{tab:ai-examples}
\end{table*}


\subsection{Semantic Clustering}
\label{app:data-cluster}

\paragraph{Embedding} 

After obtaining the six key aspects of the papers, we first leverage the InstructionEmbedding model (instructor large)\cite{su2022one} to obtain the semantic embeddings of the scientific problems and AI methods extracted.
InstructorEmbedding is an advanced embedding model that generates task- and domain-specific text embeddings without additional training. Given our need to embed both scientific problems and AI methods, we use InstructorEmbedding with instructions below to provide contextualized embeddings tailored to our task rather than context-free embedding methods

For scientific problems, we embed the problem keywords/keyphrases concatenated with the problem definition with the following instructions:

\begin{framed}
\noindent Represent the keyphrase and definition of a scientific problem for clustering and visualizing scientific problems
\end{framed}

For AI methods, we embed the AI method keywords/keyphrases concatenated with the AI method definition with the following instructions: 

\begin{framed}
\noindent Represent the Artificial Intelligence method paragraph for clustering and visualizing Artificial Intelligence methods
\end{framed}

\paragraph{Clustering}

After we obtained the semantic embeddings, we used LargeVis \cite{tang2016visualizing} for dimension reduction followed by HDBSCAN \cite{campello2013density} for clustering.
HDBSCAN is a widely used density-based clustering algorithm designed for practical data mining tasks. As an extension of DBSCAN, it employs hierarchical clustering to identify clusters of varying densities and is more robust to parameter selection. A density-based clustering method allows us to identify clusters at appropriate granularities and avoids a top-down decision to predefine the number of clusters. Its efficiency ($O(n\log n)$ complexity) also makes it suitable for our dataset size.

\paragraph{Labeling} \label{app:GPT_summary}

To have more consistent and generalizable knowledge of the clusters, we asked GPT (\texttt{gpt-4o-2024-08-06}) to summarize each cluster. We first perform TF-IDF on each of the clusters, obtaining a list of top words. Then we provide the top words as well as examples of paper key aspects from a certain cluster to the LLM for summarization.

Using LLMs with the below prompts to summarize clusters encourage the cluster labels to have the desirable properties \cite{mei2007automatic}:
(1) understandable, (2) semantically relevant, (3) well covering the whole topic, and (4) distinguishable from other topics.

For scientific problem clusters we use the following prompt: 

\begin{framed}
\noindent\#\# Background and Task Description\newline
You are an assistant in scientific research. Given a cluster of scientific problems/challenges, please help summarizing the cluster into a keyword or keyphrase less than 3 words. Moreover, top words from several nearby clusters are provided and your summary should accurately capture the essence of the cluster while differentiating it from neighboring clusters. Please summarize based on the provided information. \newline \newline
* Problem (keyword/keyphrase): A keyword or a keyphrase that summarizes the main problem to be addressed in this paper. \newline
* Problem (definition): The detailed definition of the problem.\newline
* Problem discipline: The discipline in which the main problem best fits. \newline \newline
\#\# Top words from texual information \newline
Below are some top words in we extracted from texual information of this cluster using TF-IDF \newline
\{ top words \} \newline \newline
\#\# Examples of Scientific Problems \newline
Below are some specific examples of scientific problems in this cluster. For each sample the problem keyword, definition and discipline will be provided. \newline
\{ examples \} \newline \newline
\#\# Top words from nearby clusters' texual information \newline
\{ Nearby clusters' top words \} \newline \newline
\#\# Summarization Requirement \newline
Based on the above information, please find a keyword or a keyphrase less than *three* words to summarize this cluster of scientific problems, satisfying the following requirements: \newline
* The summarization should be able to cover all the given top words and examples (i.e., with a high "recall")\newline
* The summarization should be specific enough so that it only covers contents in this cluster (i.e., with a high "precision")\newline
* If the summarization is not possible due to lack of information, output "N/A" when applicable.\newline
* The summarization should be specific enough to differentiate it from nearby clusters\newline
* Also provide summarization for nearby clusters\newline \newline
\#\# Response Format \newline
Please output the summarization as a list, which has exactly *(\#nearby clusters+1) outputs. The output should be in the format of a python list. \newline
["Keyword", "Keyword neighbor 1",...]
\end{framed}

For AI method clusters we use the following prompt: 

\begin{framed}
\noindent\#\# Background and Task Description \newline
You are an assistant in Artificial Intelligence research, where where AI generally refers to intelligence exhibited by machines (particularly computer systems), including models, algorithms. Given a cluster of Artificial Intelligence methods, please help summarizing the cluster into a keyword or keyphrase less than 3 words. The top appearing words in the cluster together with some examples of Artificial Intelligence methods from that cluster will be provided. Moreover, top words from several nearby clusters are provided and your summary should accurately capture the essence of the cluster while differentiating it from neighboring clusters.. Please summarize based on the provided information. \newline
* Method (keyword/keyphrase): A keyword or a keyphrase that summarizes the main method used in this paper to address the above problem. \newline
* Method (definition): The detailed definition of the method. \newline \newline
\#\# Top words from texual information \newline
Below are some top words in we extracted from texual information of this cluster using TF-IDF \newline
\{ top words \} \newline \newline
\#\# Examples of Artificial Intelligence methods \newline
Below are some specific examples of Artificial Intelligence methods in this cluster. For each sample the method keyword and definition will be provided. \newline
\{examples\} \newline \newline
\#\# Top words from nearby clusters' texual information \newline
\{ Nearby clusters' top words \} \newline \newline
\#\# Summarization Requirement \newline
Based on the above information, please find a keyword or a keyphrase less than *three* words to summarize this cluster of AI methods, satisfying the following requirements: \newline
* The summarization should be able to cover all the given top words and examples (i.e., with a high "recall") \newline
* The summarization should be specific enough so that it only covers contents in this cluster (i.e., with a high "precision") \newline
* If the summarization is not possible due to lack of information, output "N/A" when applicable. \newline
* The summarization should be specific enough to differentiate it from nearby clusters \newline
* Also provide summarization for nearby clusters \newline \newline
\#\# Response Format \newline
Please output the summarization as a list, which has exactly *(\#nearby clusters)* outputs. The output should be in the format of a python list. \newline
["Keyword", "Keyword neighbor 1",...]
\end{framed}

\section{Discrepancy Analysis}

\begin{figure*}[htbp]
    \centering
    \includegraphics[width=1.0\linewidth]{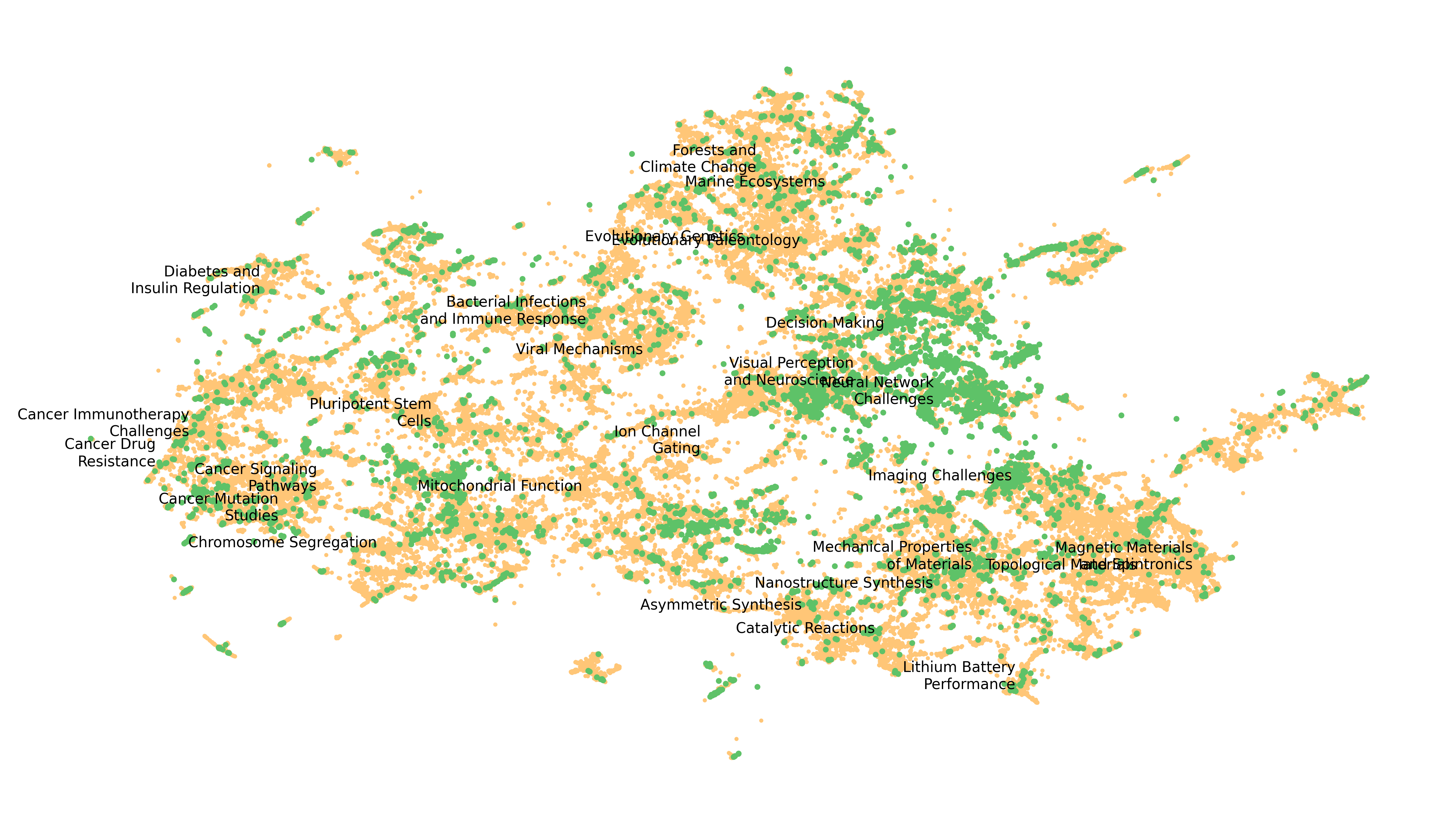}
    \caption{Distribution of scientific problems of all publications, annotated with gpt-4o-2024-08-06 summary.
    }
    \label{fig:science_all}
\end{figure*}

\begin{figure*}[htbp]
    \centering
    \includegraphics[width=1.0\linewidth]{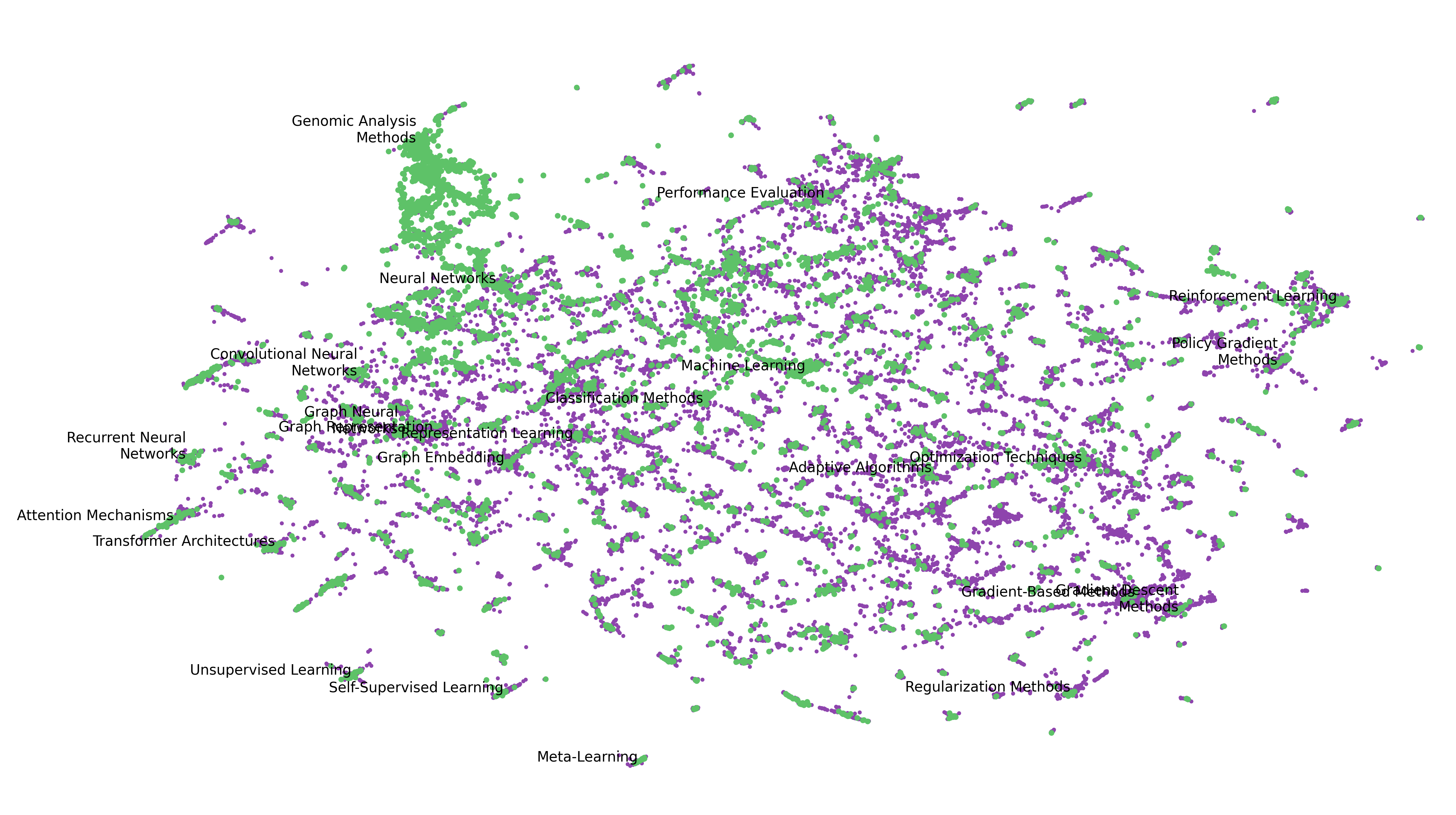}
    \caption{Distribution of AI methods of all publications, annotated with gpt-4o-2024-08-06 summary.}
    \label{fig:AI_all}
\end{figure*}

Fig. \ref{fig:science_all}-\ref{fig:AI_all} illustrates the distribution of AI4Science and non-AI4Science publications across different communities, highlighting how AI and scientific fields engage with interdisciplinary research distinctively.

\paragraph{Well- and under-explored regions}

Table \ref{tab:top-clusters-explored} list some top clusters of well- and under-explored scientific problems as well as AI methods. 

\begin{table*}[htbp]
    \centering
    \small
    \begin{tabular}{|l|l|l|l|}
    \hline
        \multicolumn{2}{|c|}{\textbf{Scientific problems}}			& 	\multicolumn{2}{c|}{\textbf{AI Methods}}			\\
        \hline
\textbf{Well-explored}	&	\textbf{Under-explored}	&	\textbf{Well-explored}	&	\textbf{Under-explored}	\\
\hline
Advanced Imaging Challenges	&	Asymmetric Synthesis	&	Deep Learning Models	&	Attention Mechanisms	\\
Visual Perception and Neuroscience	&	Magnetic Materials and Spintronics	&	Convolutional Neural Networks	&	Gradient Descent Methods	\\
Decision Making	&	Forests and Climate Change	&	Genomic Analysis	&	Gradient-Based Methods	\\
Neural Network Challenges	&	Viral Mechanisms	&	Neural Networks	&	Policy Gradient Methods	\\
Urban Traffic Management	&	Evolutionary Paleontology	&	Classification Methods	&	Graph Embedding	\\
Alzheimer's Disease Mechanisms	&	Nanostructure Synthesis	&	Computational Modeling	&	Regularization Methods	\\
Statistical Inference	&	Diabetes and Insulin Regulation	&	Molecular Dynamics Simulations	&	Transformer Architectures	\\
Depression and Anxiety	&	Bacterial Infections and Immune Response	&	Predictive Modeling	&	Contrastive Learning	\\
Language Processing	&	Cancer Signaling Pathways	&	Advanced Imaging Techniques	&	Graph Representation	\\
Genetic Variants and Traits	&	Cancer Drug Resistance	&	Computational Drug Screening	&	Adaptive Algorithms	\\
\hline
    \end{tabular}
    \caption{Scientific problems and AI methods that are well- and under-explored by AI4Science research. Cluster names are ranked by the number of total publications. Only top 10 clusters are shown. }
    \vspace{-10pt}
    \label{tab:top-clusters-explored}
\end{table*}

\paragraph{Bipartite graph node degrees. }
Table \ref{tab:degree-highest} presents the nodes with the highest degrees in the bipartite graph, representing the most connected scientific challenges and AI methods in the AI4Science landscape. These high-degree nodes act as key ``hubs,'' indicating their central role in bridging AI techniques with scientific problems. 

\paragraph{Discrepancies between the AI and science communities. }

\begin{figure*}[htbp]
    \centering
    \begin{subfigure}[htbp]{0.48\linewidth}
        \centering
        \includegraphics[width=\linewidth]{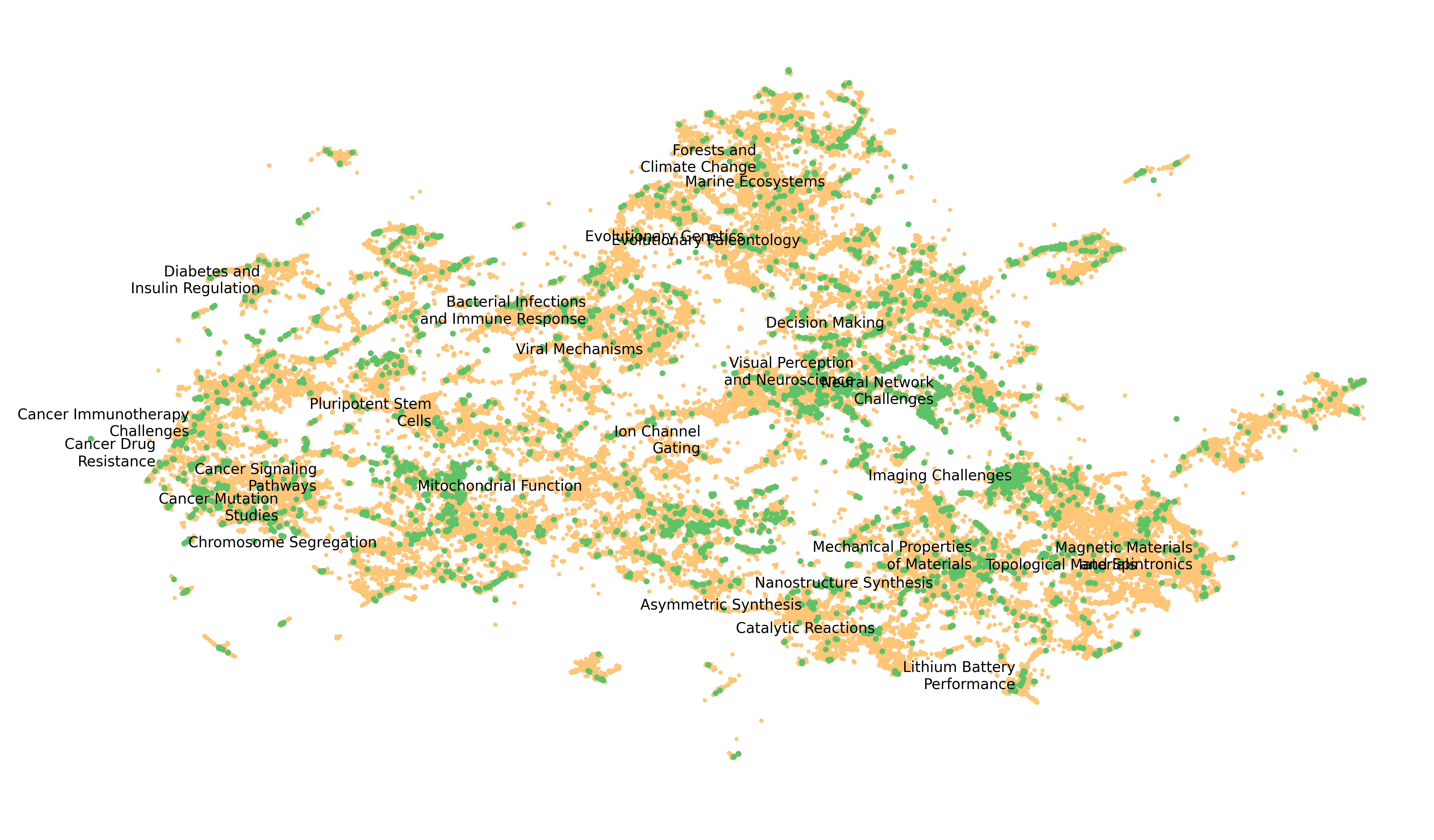}
        \caption{Distribution of scientific problems of scientific journal papers, annotated with gpt-4o-2024-08-06 summary.}
        \label{fig:science_sci}
    \end{subfigure}
    \hfill
    \begin{subfigure}[htbp]{0.48\linewidth}
        \centering
        \includegraphics[width=\linewidth]{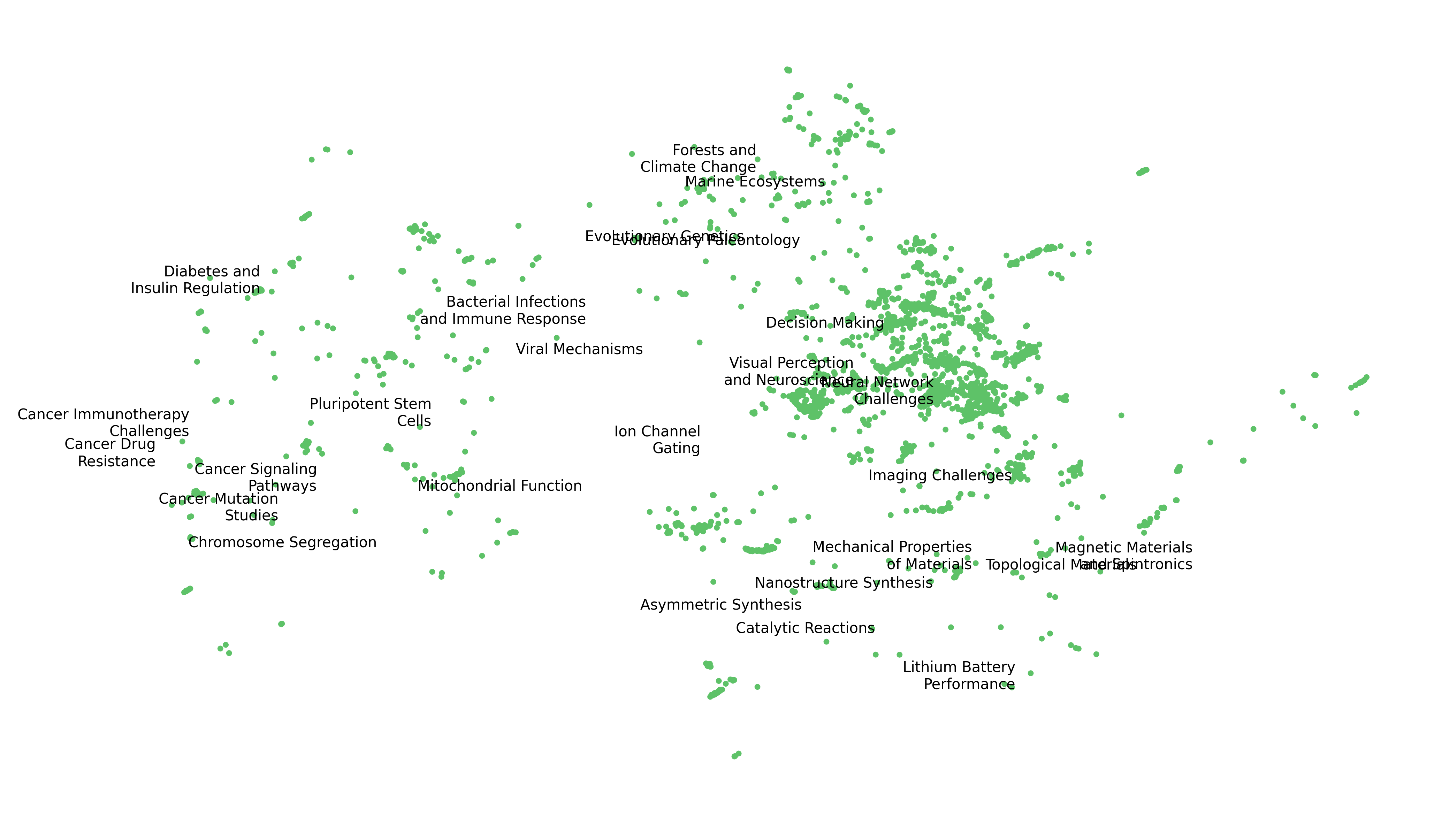}
        \caption{Distribution of scientific problems of AI conference papers, annotated with gpt-4o-2024-08-06 summary.}
        \label{fig:science_cs}
    \end{subfigure}

    \begin{subfigure}[htbp]{0.48\linewidth}
        \centering
        \includegraphics[width=\linewidth]{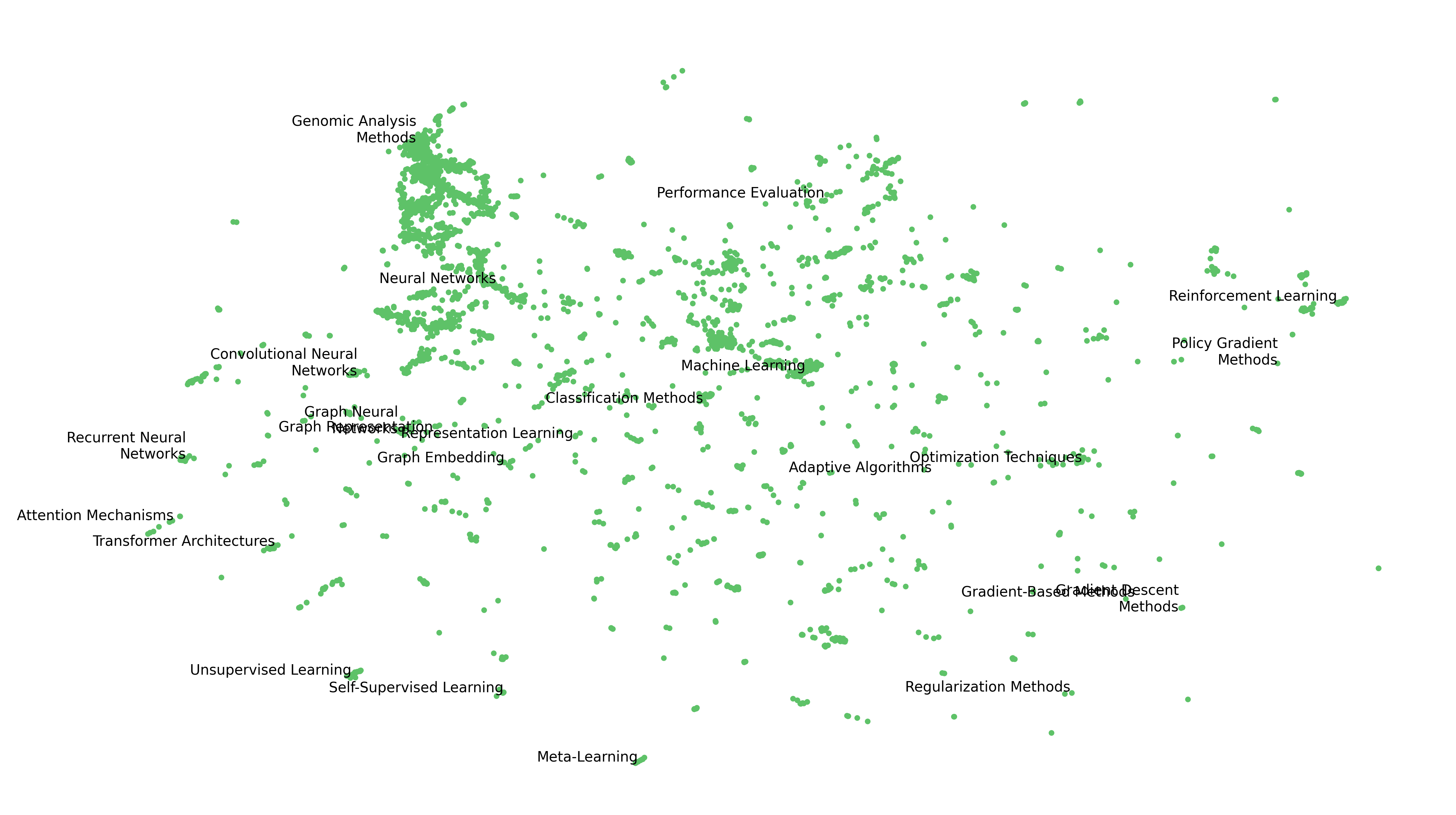}
        \caption{Distribution of AI methods of scientific journal papers, annotated with gpt-4o-2024-08-06 summary.}
        \label{fig:AI_Sci}
    \end{subfigure}
    \hfill
    \begin{subfigure}[htbp]{0.48\linewidth}
        \centering
        \includegraphics[width=\linewidth]{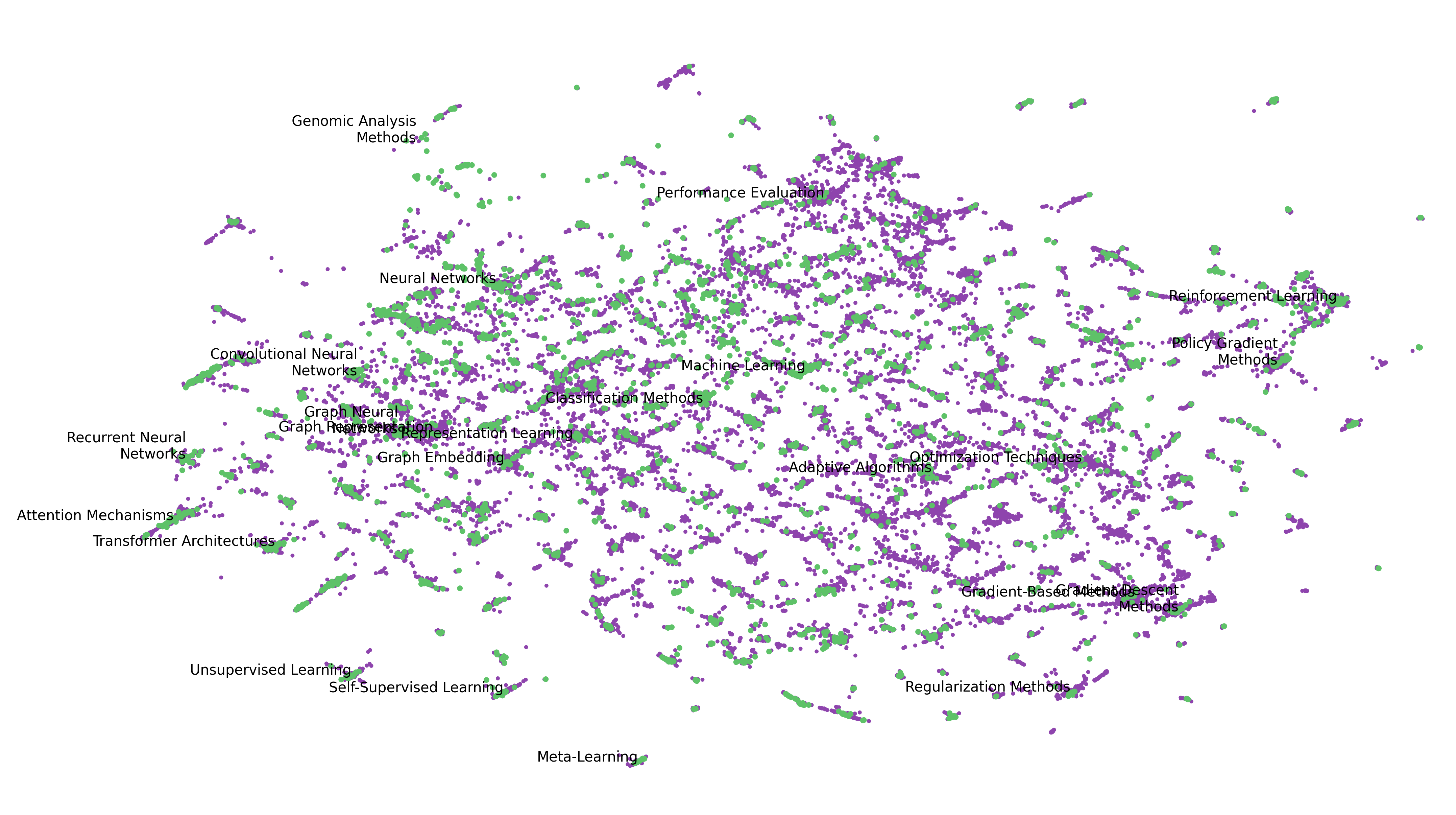}
        \caption{Distribution of AI methods of AI conference papers, annotated with gpt-4o-2024-08-06 summary.}
        \label{fig:AI_CS}
    \end{subfigure}

    \caption{Distribution of scientific problems and AI methods, annotated with gpt-4o-2024-08-06 summary.}
    \label{fig:combined_distribution}
\end{figure*}

Table \ref{tab:communities} lists the top clusters emphasized by each community, revealing distinct focus areas and approaches in integrating AI into scientific research.

\begin{table*}[]
\centering
\begin{tabular}{|r|lr|lr|}
\hline 
& \textbf{Scientific Problem Node}                       & \textbf{Deg.} & \textbf{AI Method Node}                     & \textbf{Deg.} \\ \hline
1                    & Neural Network Challenges           & 164                               & Machine Learning                   & 140                               \\
2                    & Urban Traffic Management               & 122                               & Genomic Analysis Methods                      & 137                               \\
3                    & Electronic Health Data Challenges    & 102                               &  Advanced Imaging Techniques                   & 98                               \\
4                    & Imaging Challenges              & 89                               & Computational Drug Screening  & 97                                \\
5                    &Statistical Inference              & 89                               & Molecular Dynamics Simulations              & 88                                \\
6                    & Decision Making          & 86                               & 
 Protein Structure Design                    & 88                                \\
7                    & Language Processing                    & 85                              & Deep Learning Models                    & 84                                \\
8                    & Dynamical Systems                         & 84                               & Neural Networks                 & 62                                \\
9                    & Visual Perception and Neuroscience                   & 82                             & Classification Methods                   & 62                               \\
10                   & Market Pricing and Allocation                 & 77                              & Molecular Structural Design             & 61                                \\
11                   & Neural Imaging Challenges                    & 74                             & Deep Neural Networks                      & 57                                \\
12                   & Single-cell RNA Sequencing         & 71                                & Predictive Modeling                 & 53                                \\
13                   & Soft Robotics                   & 69                                & Convolutional Neural Networks              & 52                               \\
14                   & Drug Discovery & 67                                & 
Computational Modeling              & 52                                \\
15                   & Protein Design and Function Prediction  & 67                                & Unsupervised Learning         & 50                               \\
16                   & Social Media Dynamics            & 67                                & Image Segmentation                & 50                              \\
17                   & Mechanical Properties of Materials        & 64                                & Data Analysis                    & 41                                \\
18                   & Causal Inference                     & 62                                & Deep Learning Framework               & 39                               \\
19                   & Optimization Problems                  & 60                                & Data-driven Approaches      & 39                                \\
20                   &Game Theory Challenges        & 58                                & Recurrent Neural Networks & 37                           
\\ \hline
\end{tabular}
\caption{Nodes in the bipartite graph (Fig. \ref{fig:bipartite}a) with the highest degrees. }
\label{tab:degree-highest}
\end{table*}

\begin{table*}[]
    \centering
    \footnotesize
    \begin{tabular}{|r|l|l|l|l|}
\hline
\multicolumn{1}{|l|}{\multirow{2}{*}{}} & \multicolumn{2}{c|}{\textbf{Top scientific problems in AI4Science research}} & \multicolumn{2}{c|}{\textbf{Top AI methods in AI4Science research}} \\
\cline{2-5}
\multicolumn{1}{|l|}{}                  & \textbf{In the science community}           & \textbf{In the AI community}               & \textbf{In the science community}                   & \textbf{In the AI community}                   \\ 
\hline
1                                     & Neural Network Challenges                     & Urban Traffic Management                    & Machine Learning                                      & Deep Learning Models                                    \\
2                                     & Imaging Challenges                        & Neural Network Challenges                        & Genomic Analysis Methods                                        & Machine Learning                                 \\
3                                     & Single-cell RNA Sequencing              & Electronic Health Data Challenges                        & Deep Learning Models                                     & Reinforcement Learning                                    \\
4                                     & Mechanical Properties of Materials                 & Language Processing             & Molecular Dynamics Simulations                     & Attention Mechanisms                                  \\
5                                     & Visual Perception and Neuroscience                  & Statistical Inference                                  & Advanced Imaging Techniques                                        & Convolutional Neural Networks                                      \\
6                                     & Protein Design and Function Prediction                              & Market Pricing and Allocation                             & Computational Drug Screening                                       & Neural Networks                               \\
7                                     & Cancer Detection                & Soft Robotics                        & Protein Structure Design                                 & Recurrent Neural Networks                                 \\
8                                     & Genetic Variants and Traits                         & Dynamical Systems                    & Molecular Structural Design                                   & Graph Neural Networks                           \\
9                                     & Decision Making                    & Decision Making                            & Brain Imaging Techniques                                     & Policy Gradient Methods                           \\
10                                    & Language Processing                             & Game Theory Challenges                             & Computational Modeling                                 & Multitask Learning                             \\
11                                    & Gene Regulatory Networks                     & Causal Inference                            & Optical AI Methods                                      & Classification Methods                              \\
12                                    & Drug Discovery                             & Drug Discovery                           & Deep Neural Networks                               & Transformer Architectures                      \\
13                                    & Cancer Mutation Studies                    & Optimization Problems           & Wireless Sensor Technology                              & Spatio-Temporal Analysis                    \\
14                                    & Mitochondrial DNA Mutations                             & Social Media Dynamics                          & Neural Networks                                        & Deep Reinforcement Learning                      \\
15                                    & Protein-Protein Interactions                             & Imaging Challenges          & Image Segmentation                            & Predictive Modeling                         \\
16                                    & Neural Imaging Challenges                    & Urban Dynamics                     & Unsupervised Learning                                 & Causal Inference                              \\
17                                    & Cryo-electron microscopy                  & Neural Imaging Challenges                         & Classification Methods                                   & Self-Supervised Learning                              \\
18                                    &Tactile Sensing Technology                     & Visual Perception and Neuroscience                & Convolutional Neural Networks                               & Graph Convolutional Networks                         \\
19                                    & Chemical Reaction Dynamics                         & Sparse Signal Processing                & Predictive Modeling                                 & Unsupervised Learning                           \\
20                                    & Chromatin Organization                    & Energy Management                     & Neuromorphic Devices                                    & Latent Variable Models                             
\\ \hline
\end{tabular}
    \caption{
    Top scientific problem and AI method clusters in AI4Science research, identified within both the science and AI communities. 
    The clusters are ranked based on the number of AI4Science publications contributed by each community.
    }
    \label{tab:communities}
\end{table*}

\section{Experiment Details}
\label{app:experiment}

\subsection{Experiment Setups}

\paragraph{Data. }

Table \ref{tab:train-test-data} lists the statistics of the training and test data splits. 

\begin{table}[htbp]
    \centering
    \small
    \begin{tabular}{c|l|rr}
        \toprule
        \multicolumn{2}{c|}{} & \textbf{Train} & \textbf{Test} \\
        \midrule 
        & Publication year range & 2014-2022 & 2023-2024 \\
        \hline
        \parbox[htbp]{2mm}{\multirow{6}{*}{\rotatebox[origin=c]{90}{Number of}}}
         & All publications & 141,639 & 21,017 \\
         & AI4Science publications & 6,287 & 1,225 \\ 
        \cline{2-4}
         & Pub. in well-explored sci. clusters & 4,262 & 594 \\
         & Pub. in under-explored sci. clusters & 1,180& 392\\ 
         \cline{2-4}
         & Pub. in well-explored AI clusters & 3,076 & 742\\
         & Pub. in under-explored AI clusters &1,387 & 198\\
        \hline
        \parbox[htbp]{2mm}{\multirow{6}{*}{\rotatebox[origin=c]{90}{Avg. deg. of}}}
         & Scientific problem nodes &11.3 & 2.8\\
         & Well-explored sci. nodes & 37.7 & 6.9 \\
         & Under-explored sci. nodes &3.9 & 1.4\\
        \cline{2-4}
         & AI method nodes & 12.4 & 3.0\\
         & Well-explored AI nodes &16.2 & 4.9\\
         & Under-explored AI nodes & 7.2 & 1.3\\
        \bottomrule
    \end{tabular}
    \caption{The training and test splits of data. 
    }
    \vspace{-20pt}
    \label{tab:train-test-data}
\end{table}

\paragraph{Models. }

We examine two approaches to link prediction based on different data representations:

\begin{itemize}
    \item \emph{Cluster-level link prediction} (``$P_i\to M_i$'' or ``$M_i\to P_i$''): In this approach, scientific problems and AI methods are represented as high-level cluster labels, serving as nodes in a bipartite graph. This captures more generalized connections between similar scientific problems and AI methods.
    \item \emph{Paper-level link prediction} (``$p_i\to m_i,u_i$'' or ``$m_i\to p_i,u_i$''): Here, scientific problems and AI methods are expressed in detailed text descriptions rather than coarse cluster labels. This approach leverages textual generative models, such as large language models (LLMs), to capture more fine-grained distinctions and predict links based on textual data.
\end{itemize}

Following \citet{ozer2024link}, we employ three categories of bipartite graph link prediction models: (1) link score-based predictions, where we utilize the Katz index calculated using common neighbors \cite{katz1953new}; 
and (2) embedding-based link predictions \cite{perozzi2014deepwalk,tang2015line}, in which we apply spectral embedding and node2vec embedding techniques \cite{grover2016node2vec}.

The Katz index between two nodes $x$ and $y$ are calculated as below:

\begin{equation}
    \text{Katz}(x,y) = \sum_\emph{l=1}^{\infty} \alpha^l(A^l)_{xy},
\end{equation}

where $A$ is the connectivity matrix and $\alpha$ is a penalty factor for path length (In all of our experiments we chose $\alpha = 0.1$). 


For generative link prediction, we employ two OpenAI GPT models, \texttt{gpt-3.5-turbo-0125} and \texttt{gpt-4o-2024-08-06}, which represent different levels of generative capabilities. It's important to note that the training data for \texttt{gpt-3.5-turbo-0125} extends only up to September 2021, meaning there is no risk of data leakage based on our training/test data splitting strategy. However, the training data for \texttt{gpt-4o-2024-08-06} includes information up to October 2023, which may introduce potential data leakage and could lead to artificially higher performance. We will discuss the impact of this potential data leakage in the results section.


\paragraph{Evaluation metrics.}

For cluster-level link prediction, we evaluate performance using Recall, Precision, and F1 scores @K, where K represents the number of generated candidate links, as is standard in the literature \cite{al2006link, yang2015evaluating}.
Particularly, for a source node $s$, supposing in the test data, it is connected to a set of target nodes $T(s)$, and the model generates $K$ predictions $P(s)$. The metrics are calculated as below:

\begin{align}
    \text{Precision} &:= \frac{\vert\{p\in T(s)\mid p\in P(s)\}\vert}{K},\\
    \text{Recall} &:= \frac{\vert\{t\in P(s)\mid t\in T(s)\}\vert}{\vert T(s)\vert}. 
\end{align}


For paper-level link prediction, we adopt three widely used evaluation metrics for text generation: ROUGE \cite{lin2004rouge}, BLEURT \cite{sellam2020bleurt}, and the cosine similarity of text embeddings. The best scores @K are reported for each of these metrics.

For ROUGE, we report rouge-1-f, which is defined as follows (given a reference's unigrams r and extraction's unigrams s): 
\begin{align*}
    p &= \frac{|s \cap r|}{|s|} \\
    r &= \frac{|s \cap r|}{|r|} \\
    f &= \frac{2*p*r}{p+r}
\end{align*}

For BLEURT \cite{sellam2020bleurt}, we use the pre-trained checkpoint \texttt{BLEURT-20}

For instruction embedding, we first calculate reference's and extraction's embedding ($r$ and $e$ respectively) via pre-trained checkpoint Instructor-large \cite{su2022one} and calculate the cosine similarity: 
$$
cos(r,e) = \frac{<r,e>}{||r|| \ ||e||}
$$

\paragraph{LLM (Paper) link prediction setup}

For paper-level generative link prediction, we consider both Sci $\to$ AI and AI $\to$ Sci generations: 

\begin{itemize}
    \item Sci $\to$ AI. Given the scientific problem description, generate the AI method that can solve the scientific problem.
    \item AI $\to$ Sci. Given the AI method description, generate the scientific problem that the AI methods can solve.
\end{itemize}

To evaluate LLM link generation quality, we consider the following experiments: \newline
\begin{itemize}
    \item Imitation @ K: The nearest K descriptions (scientific problems or AI methods) are imitated to be the predictions. Distance is measured by semantic embedding cosine similarity (Appendix \ref{app:data-cluster}). 
    \item Direct @ K: In this scenario, we directly ask the LLM to generate K predictions given the extracted scientific problem/AI methods description.
    \item RAG @ K: We first find $n$ most similar examples (similarity measure by cosine similarity of semantic embedding) and provide these examples to the LLM to generate K predictions given the extracted scientific problem/AI methods descriptions. We experiment with $n=1,3,5$. In the main paper, we report the results with $n=5$.
\end{itemize}

\paragraph{Prompts}

For paper-level link prediction with LLMs, we use the following prompt\footnote{For brevity only Sci $\to$ AI RAG prompt is provided}: 

\begin{framed}
\noindent \#\# Background and Task Description \newline
You are an expert in both science and artificial intelligence (AI), where AI generally refers to intelligence exhibited by machines (particularly computer systems), including models, algorithms, etc. Given a scientific problem, your task is to recommend potential Artificial Intelligence methods that can be used to address this scientific problem. A few examples of relevant papers with similar scientific problems will be provided.\newline \newline
* Problem (keyword/keyphrase): A keyword or a keyphrase that summarizes the main problem to be addressed in this paper. \newline
* Problem (definition): The detailed definition of the problem. \newline 
* Problem discipline: The discipline in which the main problem best fits. \newline
* Method (keyword/keyphrase): A keyword or a keyphrase that summarizes the main method used in this paper to address the above problem. \newline 
* Method (definition): The detailed definition of the method. \newline
* Usage: How the method is specifically applied to address the problem. \newline \newline
\#\# Scientific Problem \newline
Please recommend an AI method to address the below scientific problem for writing an academic paper: \newline
\{Key Aspects Extraction\} \newline \newline
\#\# Examples of AI usage in similar scientific papers:
\{examples\} \newline \newline
\#\# Notes \newline
* If no potential AI method can be used, mark it as "N/A" (not applicable). \newline
* Please respond with specific AI methods instead of high-level AI methods.\newline
* Exactly output one recommendation.\newline \newline
\#\# Response Format \newline
Please output the recommendation of AI methods as a list, which has exactly *one* element. The output should be in the format of a list as below:\newline
[ \newline
\{ \newline
        "AI Method (keyword/keyphrase)": "...", \newline
        "AI Usage": "..." \newline
    \}, \newline
    ... \newline
]
    \end{framed}
For cluster-level link prediction with LLMs (LLM (Cluster)), we use the following prompt: \footnote{For brevity only Sci $\to$ AI RAG prompt is provided}: 

\begin{framed}
\noindent \#\# Background and Task Description \newline
You are an expert in both science and artificial intelligence (AI), where AI generally refers to intelligence exhibited by machines (particularly computer systems), including models, algorithms, etc. Given a scientific problem domain and past usage of Artificial Intelligence methods in solving scientific problems, your task is to recommend potential Artificial Intelligence methods that can be used to address the scientific problem. \newline
\newline
\#\# Scientific problem domain\newline
\{sci cluster\} \newline
\newline
\#\# Possible Artificial Intelligence domains\newline
\{AI clusters\}\newline
\newline
\#\# Format of past usage of AI methods to solve Scientific problems:\newline
(u,v,k): Scientific problem u has been solved with AI method v for k times in previous scientific literature.\newline
\newline
\#\# Past usage of AI methods to solve Scientific problems:\newline
\{example links\}\newline
\newline
\#\# Notes \newline
* If no potential AI method can be used, mark it as "N/A" (not applicable). \newline
* The AI methods recommended should be within the possible Artificial Intelligence domains.\newline
* The AI methods recommended may or may not be within the given observed links.\newline
* Exactly recommend \{k\} AI methods.\newline
\newline
\#\# Response Format \newline
please output the recommended AI methods as a list, which contains exactly \{k\} elements. The output should be in the list format as below:\newline
\newline
[\newline
    "Artificial Intelligence Method 1",\newline
    ...\newline
]\newline
    \end{framed}

\subsection{Experiment Results}

Tables \ref{tab:results-app-beg}-\ref{tab:results-app-end} summarize the detailed LLM link generation experiment results with different models and evaluation metrics. These results show that: 
\begin{itemize}
    \item Including RAG examples helps the model make more informed predictions, as all evaluation metric results improve when we increase the examples given from $n=1$ to $n=5$.
    \item While Imitation can achieve higher rouge scores, it has less semantic similarity compared with LLM generations.
    \item Table \ref{tab:model_comparison} show that the performance of LLM (Paper) remains consistent across different models used, suggesting a low risk of data leakage. However, the performance of LLM (Cluster) declines, likely due to GPT-3.5’s limited ability of reasoning and processing long contexts. 
\end{itemize}

\begin{table*}[htbp]
\small
\centering
\begin{tabular}{|l|l|cccc|cccc|cccc|}
\hline
\textbf{Setting} & \textbf{Model} & \multicolumn{4}{c|}{\textbf{Precision}} & \multicolumn{4}{c|}{\textbf{Recall}} & \multicolumn{4}{c|}{\textbf{F1}} \\
& & @1 & @3 & @5 & @10 & @1 & @3 & @5 & @10 & @1 & @3 & @5 & @10 \\ \hline
\multirow{4}{*}{Sci $\to$ AI} & LLM (Paper,gpt-3.5-turbo-0125) & 0.273 &	0.191 &	0.175 &	0.141 & 0.052 &	0.081 &	0.109 &	0.134 & 0.087 &	0.114 &	0.134 &	0.138 \\ 
 & LLM (Paper,gpt-4o-2024-08-06) & 0.282 &	0.227 &	0.200 &	0.156 & 0.058 &	0.106 &	0.129 &	0.155 & 0.096 &	0.145 &	0.157 &	0.155  \\
 & LLM (Cluster,gpt-3.5-turbo-0125) & 0.134 & 0.168 &	0.167 &	0.137 &	0.014 &	0.065 &	0.108 &	0.180 &	0.025 &	0.094 &	0.131 &	0.155  \\
& LLM (Cluster,gpt-4o-2024-08-06) & 0.352 & 0.300 &	0.238 &	0.159 &  0.053 &	0.101 &	0.134 &	0.176 & 0.093 &	0.151 &	0.171 &	0.167  \\ 
\hline
\multirow{4}{*}{AI $\to$ Sci} & LLM (Paper,gpt-3.5-turbo-0125) & 0.516 &	0.421 &	0.401 &	0.210 & 0.099 &	0.127 &	0.148 & 0.148 &	0.166 &	0.195 &	0.216 &	0.174 \\
 & LLM (Paper,gpt-4o-2024-08-06) & 0.377 &	0.323 &	0.307 &	0.296 & 0.065 &	0.104 &	0.127 &	0.163 & 0.111 &	0.158 &	0.180 &	0.211 \\
& LLM (Cluster,gpt-3.5-turbo-0125)  & 0.136 &	0.082 &	0.120 &	0.102 & 0.009 &	0.023 &	0.053 &	0.088 & 0.017 &	0.036 &	0.074 &	0.095 \\ 
 & LLM (Cluster,gpt-4o-2024-08-06)  & 0.201 &	0.168 &	0.143 &	0.129 & 0.016 &	0.042 &	0.053 &	0.092& 0.030 &	0.068 &	0.077 &	0.107 \\ 
\hline
\end{tabular}
\caption{Link prediction results comparison between gpt-3.5-turbo-012 and gpt-4o-2024-08-06.
}
\label{tab:model_comparison}
\end{table*}

\begin{table*}[htbp]
    \centering
    \begin{tabular}{|l|c|c|c|c|c|c|c|c|c|c|c|c|}
    \hline
    \multicolumn{1}{|c|}{\textbf{Sci $\to$ AI}} & \multicolumn{4}{c|}{\textbf{ROUGE-1-F}} & \multicolumn{4}{c|}{\textbf{BLEURT}} &  \multicolumn{4}{c|}{\textbf{CosineSim}} \\ 
    \hline
    & \textbf{@1} & \textbf{@3} & \textbf{@5} & \textbf{@10} &  \textbf{@1} & \textbf{@3} & \textbf{@5} & \textbf{@10} & \textbf{@1} & \textbf{@3} & \textbf{@5} & \textbf{@10} \\
    \hline 
    \multirow{1}{*}Imitation  & 0.254 & \textbf{0.297} & \textbf{0.313} & \textbf{0.330} & 0.365 & 0.397 & 0.407 & 0.420 & 0.853 & 0.871 & 0.877 & 0.883 \\
    \hline
    \multirow{1}{*}{Direct} & 0.263 & 0.286 & 0.288 & 0.292 & 0.407 & 0.428 & 0.429 & 0.429 & \textbf{0.889} & \textbf{0.895} & 0.895 & 0.896 \\
    \hline
    RAG n = 1  & 0.276 & 0.291 & 0.300 & 0.309 & 0.416 & \textbf{0.434} & 0.439 & \textbf{0.444} & 0.888 & \textbf{0.895} & \textbf{0.898} & \textbf{0.901} \\
    RAG n = 3 & 0.286 & 0.296 & 0.304 & 0.313 & \textbf{0.417} & \textbf{0.434} & 0.439 & 0.443 & 0.888 & \textbf{0.895} & 0.897 & \textbf{0.901} \\
    RAG n = 5 & \textbf{0.290} & \textbf{0.297} & 0.305 & 0.314 & \textbf{0.417} & 0.432 & \textbf{0.440} & \textbf{0.444} & 0.887 & 0.894 & 0.897 & 0.900 \\
    \hline
    \end{tabular}
    \caption{gpt-3.5-turbo-0125 AI method link generation results}
    \label{tab:results-app-beg}
\end{table*}

\begin{table*}[htbp]
    \centering
    \begin{tabular}{|l|c|c|c|c|c|c|c|c|c|c|c|c|}
    \hline
    \multicolumn{1}{|c|}{\textbf{Sci $\to$ AI}} & \multicolumn{4}{c|}{\textbf{ROUGE-1-F}} & \multicolumn{4}{c|}{\textbf{BLEURT}} &  \multicolumn{4}{c|}{\textbf{CosineSim}} \\ 
    \hline
    & \textbf{@1} & \textbf{@3} & \textbf{@5} & \textbf{@10} &  \textbf{@1} & \textbf{@3} & \textbf{@5} & \textbf{@10} & \textbf{@1} & \textbf{@3} & \textbf{@5} & \textbf{@10} \\
    \hline 
    \multirow{1}{*}Imitation  & 0.254 & 0.297 & 0.313 & 0.330 & 0.365 & 0.397 & 0.407 & 0.420 & 0.853 & 0.871 & 0.877 & 0.883 \\
    \hline
    \multirow{1}{*}{Direct}  & 0.245 & 0.273 & 0.284 & 0.286 & 0.380 & 0.410 & 0.429 & 0.435 & 0.885 & 0.891 & 0.894 & 0.895 \\
    \hline
    \multirow{1}{*}{RAG n = 1} & 0.266 & 0.283 & 0.289 & 0.294 & 0.391 & 0.424 & 0.438 & 0.443 & 0.889 & 0.894 & 0.897 & 0.898 \\
    \multirow{1}{*}{RAG n = 3} & 0.273 & 0.292 & 0.295 & 0.300 & 0.395 & 0.426 & 0.441 & 0.447 & 0.890 & 0.895 & 0.898 & 0.899 \\
    \multirow{1}{*}{RAG n = 5} & 0.277 & 0.296 & 0.300 & 0.304 & 0.397 & 0.430 & 0.442 & 0.449 & 0.890 & 0.896 & 0.898 & 0.900 \\
    \hline
    \end{tabular}
    \caption{gpt-4o-2024-08-06 AI method link generation results}
\end{table*}

\begin{table*}[htbp]
    \centering
    \begin{tabular}{|l|c|c|c|c|c|c|c|c|c|c|c|c|}
    \hline
    \multicolumn{1}{|c|}{\textbf{AI $\to$ Sci}} & \multicolumn{4}{c|}{\textbf{ROUGE-1-F}} & \multicolumn{4}{c|}{\textbf{BLEURT}} &  \multicolumn{4}{c|}{\textbf{CosineSim}} \\ 
    \hline
    & \textbf{@1} & \textbf{@3} & \textbf{@5} & \textbf{@10} &  \textbf{@1} & \textbf{@3} & \textbf{@5} & \textbf{@10} & \textbf{@1} & \textbf{@3} & \textbf{@5} & \textbf{@10} \\
    \hline 
    \multirow{1}{*}Imitation & 0.241 & 0.281 & 0.293 & 0.310 & 0.342 & 0.372 & 0.379 & 0.389 & 0.795 & 0.819 & 0.826 & 0.836 \\
    \hline
    \multirow{1}{*}{Direct}  & 0.221 & 0.236 & 0.242 & 0.255 & 0.383 & 0.394 & 0.397 & 0.403 & 0.822 & 0.834 & 0.837 & 0.846 \\
    \hline
    RAG n = 1 & 0.248 & 0.249 & 0.253 & 0.265 & 0.389 & 0.390 & 0.398 & 0.402 & 0.823 & 0.832 & 0.837 & 0.845 \\
    RAG n = 3 & 0.246 & 0.257 & 0.263 & 0.261 & 0.388 & 0.399 & 0.402 & 0.404 & 0.818 & 0.832 & 0.839 & 0.842 \\
    RAG n = 5 & 0.255 & 0.267 & 0.266 & 0.276 & 0.387 & 0.395 & 0.402 & 0.407 & 0.818 & 0.829 & 0.838 & 0.845 \\
    \hline
    \end{tabular}
    \caption{gpt-3.5-turbo-0125 scientific problem link generation results}
\end{table*}

\begin{table*}[htbp]
    \centering
    \begin{tabular}{|l|c|c|c|c|c|c|c|c|c|c|c|c|}
    \hline
    \multicolumn{1}{|c|}{\textbf{AI $\to$ Sci}} & \multicolumn{4}{c|}{\textbf{ROUGE-1-F}} & \multicolumn{4}{c|}{\textbf{BLEURT}} &  \multicolumn{4}{c|}{\textbf{CosineSim}} \\ 
    \hline
    & \textbf{@1} & \textbf{@3} & \textbf{@5} & \textbf{@10} &  \textbf{@1} & \textbf{@3} & \textbf{@5} & \textbf{@10} & \textbf{@1} & \textbf{@3} & \textbf{@5} & \textbf{@10} \\
    \hline 
    \multirow{1}{*}Imitation & 0.241 & 0.281 & 0.293 & 0.310 & 0.342 & 0.372 & 0.379 & 0.389 & 0.795 & 0.819 & 0.826 & 0.836 \\
    \hline
    \multirow{1}{*}{Direct} & 0.227 & 0.251 & 0.256 & 0.254 & 0.352 & 0.379 & 0.388 & 0.392 & 0.822 & 0.840 & 0.845 & 0.850 \\
    \hline
    RAG n = 1 & 0.248 & 0.267 & 0.274 & 0.287 & 0.356 & 0.386 & 0.392 & 0.406 & 0.820 & 0.839 & 0.845 & 0.853 \\
    RAG n = 3 & 0.243 & 0.272 & 0.283 & 0.288 & 0.355 & 0.385 & 0.397 & 0.403 & 0.820 & 0.836 & 0.841 & 0.851 \\
    RAG n = 5 & 0.249 & 0.279 & 0.280 & 0.285 & 0.357 & 0.390 & 0.397 & 0.400 & 0.821 & 0.838 & 0.844 & 0.845 \\
    \hline
    \end{tabular}
    \caption{gpt-4o-2024-08-06 scientific problem link generation results}
\end{table*}

\begin{table*}[]
\centering
\begin{tabular}{|l|l|cccc|cccc|cccc|}
\hline
\multicolumn{2}{|c|}{\textbf{Link prediction setting}} & \multicolumn{4}{c|}{\textbf{ROUGE-1-F}} & \multicolumn{4}{c|}{\textbf{BLEURT}} & \multicolumn{4}{c|}{\textbf{CosineSim}} \\
\multicolumn{2}{|c|}{} & @1 & @3 & @5 & @10 & @1 & @3 & @5 & @10 & @1 & @3 & @5 & @10 \\ \hline
\multirow{3}{*}{Sci $\to$ AI} & All test data & 0.290 & 0.297 & 0.305 & 0.314 & 0.417 & 0.432 & 0.440 & 0.444 & 0.887 & 0.894 & 0.897 & 0.900 \\ 
 & Well-explored & 0.284 & 0.300 & 0.308 & 0.317 & 0.418 & 0.439 & 0.446 & 0.447 & 0.882 & 0.890 & 0.893 & 0.885 \\
 & Under-explored & 0.301 & 0.316 & 0.324 & \textbf{0.333} & 0.429 & 0.451 & 0.459 & \textbf{0.460} & 0.896 & 0.904 & \textbf{0.906} & 0.902 \\ \hline
\multirow{3}{*}{AI $\to$ Sci} & All test data & 0.255 & 0.267 & 0.266 & \textbf{0.276} & 0.387 & 0.395 & 0.402 & 0.407 & 0.818 & 0.829 & 0.838 & 0.845 \\
 & Well-explored & 0.243 & 0.251 & 0.258 & 0.262 & 0.383 & 0.389 & 0.395 & 0.399 & 0.827 & 0.840 & 0.848 & \textbf{0.852} \\
 & Under-explored & 0.246 & 0.255 & 0.265 & 0.264 & 0.390 & 0.398 & 0.406 & \textbf{0.410} & 0.822 & 0.833 & 0.844 & 0.848 \\ \hline
\end{tabular}
\caption{LLM+RAG performances on text generation metrics. 
}
\vspace{-20pt}
\label{tab:results-app-end}
\end{table*}

\section{Supplementary Materials}

We include our dataset, code, analyses, and experiment results in the anonymous repository(\url{https://github.com/charles-pyj/Bridging-AI-and-Science}) as supplementary materials to this paper, which include:

\begin{itemize}
    \item The AI4Science dataset with publication extractions and cluster labels.
    \item Tables of cluster statistics for generating Fig. \ref{fig:cluster-sizes}.
    \item Link data of the bipartite graph and new links introduced by LLM-based models.
    \item Degree information of scientific problem nodes and AI method nodes.
    \item Human annotations and evaluation results with two human annotators.
    \item Code implementations for visualizations, link predictions, metrics, and data analysis, which generates the main results of the paper.
\end{itemize}

\end{document}